\PassOptionsToPackage{table}{xcolor}

\documentclass[10pt,twocolumn,letterpaper]{article}

\usepackage{cvpr}              
\usepackage{cuted}
\usepackage{float}
\usepackage{algorithm}
\usepackage{algorithmic}
\usepackage{makecell}
\usepackage{multirow}
\usepackage{xcolor}
\usepackage{colortbl}
\usepackage{float}
\usepackage{algorithm}
\usepackage{algorithmic}
\usepackage{bbding}
\usepackage{array}
\usepackage{makecell} 
\usepackage{graphicx}
\usepackage{subcaption}

\usepackage{marvosym}
\definecolor{cvprblue}{rgb}{0.21,0.49,0.74}
\usepackage[pagebackref,breaklinks,colorlinks,allcolors=cvprblue]{hyperref}

\def\paperID{4797} 
\def\confName{CVPR}
\def\confYear{2025}

\title{O-DisCo-Edit: Object Distortion Control for Unified Realistic Video Editing}

\author{
Yuqing Chen\textsuperscript{$1,3$}\thanks{This work was done during an internship at Huawei Inc.} \quad
Junjie Wang\textsuperscript{$1$ \textsuperscript{\Letter}} \quad
Lin Liu\textsuperscript{$2$ \textsuperscript{\Letter}}\thanks{Project Leader.}\quad
Ruihang Chu\textsuperscript{$1$} \quad \\
Xiaopeng Zhang\textsuperscript{$2$} \quad
Qi Tian\textsuperscript{$2$} \quad 
Yujiu Yang\textsuperscript{$1$} \quad
\\
$^1$ Tsinghua University.\quad 
$^2$ Huawei Inc.\quad 
$^3$ Pengcheng National Laboratory.\quad 
}

\begin{document}
\maketitle

\renewcommand{\thefootnote}{\Letter}
\footnotetext[0]{Corresponding authors: \texttt{wangjunjie@sz.tsinghua.edu.cn}, \texttt{ll0825@mail.ustc.edu.cn}}
\renewcommand{\thefootnote}{\arabic{footnote}} 

\begin{strip}
    \centering
    \vspace{-5em}
    \centering
    \includegraphics[width=\textwidth]{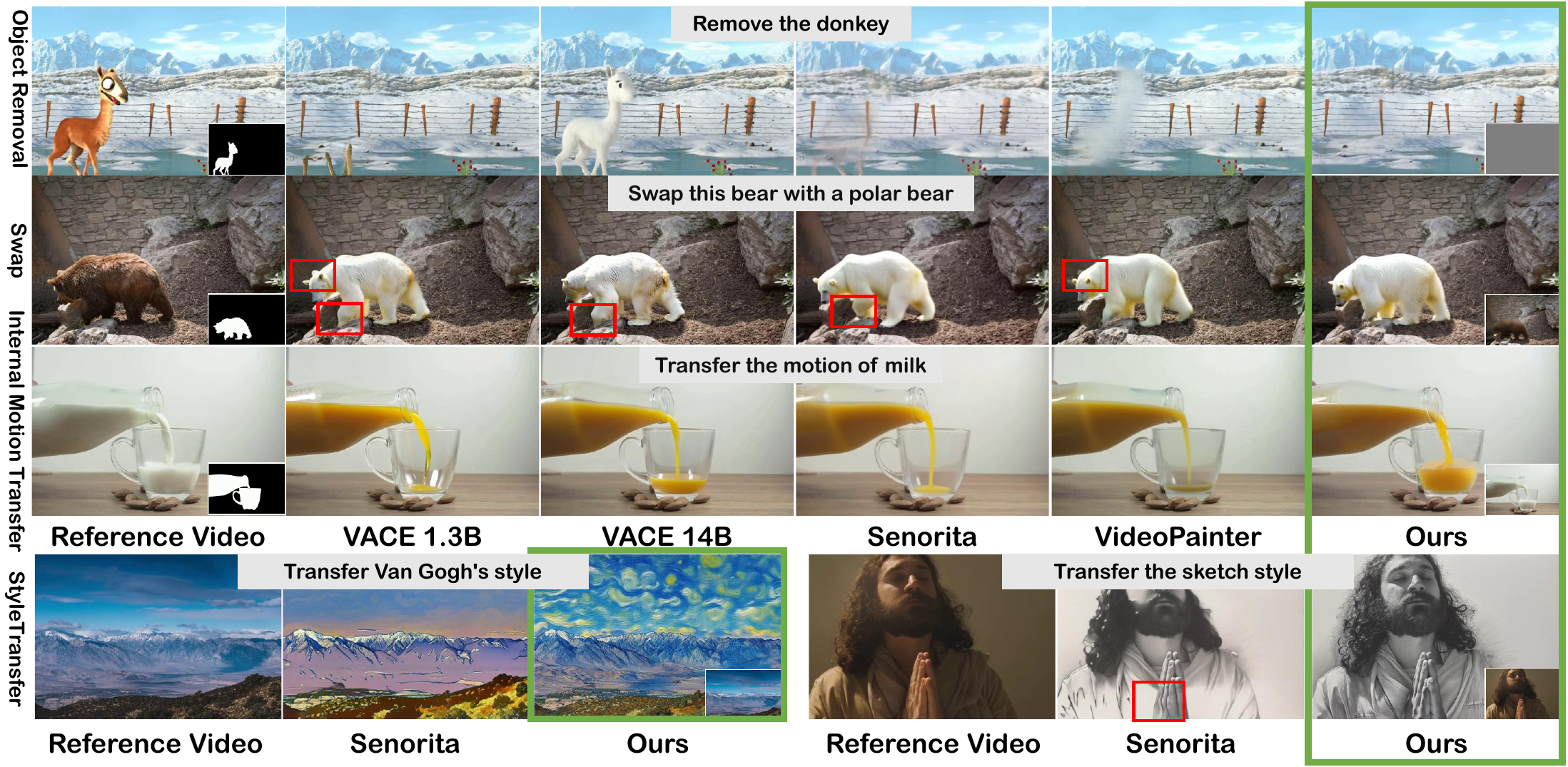}
    \captionof{figure}{Given a reference video and  image (typically the edited first frame), our method generates more realistic edited videos than SOTA approaches (VACE, Senorita and VideoPainter) across various tasks, including object removal, swap, object inside motion transfer, and style transfer. Zoom in to examine the visualization results. The bottom right of the reference video shows the input masks for all models, while the bottom right of our result displays our proposed novel control signal. 
    }
    \label{fig:remove_style_inside_swap}
\end{strip}

\begin{abstract}
Diffusion models have recently advanced video editing, yet controllable editing remains challenging due to the need for precise manipulation of diverse object properties. Current methods require  different control signal for diverse editing tasks, which complicates model design and demands significant training resources. 
To address this, we propose O-DisCo-Edit, a unified framework that incorporates a novel object distortion control (O-DisCo). This signal, based on random and adaptive noise, flexibly encapsulates a wide range of editing cues within a single representation. Paired with a ``copy-form'' preservation module for preserving non-edited regions, O-DisCo-Edit enables efficient, high-fidelity editing through an effective training paradigm. Extensive experiments and comprehensive human evaluations consistently demonstrate that O-DisCo-Edit surpasses both specialized and multitask state-of-the-art methods across various video editing tasks. \href{https://cyqii.github.io/O-DisCo-Edit.github.io/}{https://cyqii.github.io/O-DisCo-Edit.github.io/}. 
\end{abstract}    
\section{Introduction}
\label{sec:intro}

Recent years have witnessed remarkable advancements in diffusion-based video generation~\citep{yang2024cogvideox, wan2025wan, hong2022cogvideo, hacohen2024ltx}.  
Beyond pure generation, video editing has emerged as a crucial extension,  which
enables modifications to reference videos based on user instructions.  
 Specifically, effective video editing necessitates precise control over the content within edited regions, while flawlessly preserving unedited areas.

\begin{figure}[t]
    \centering
    \includegraphics[width=1.0\linewidth]{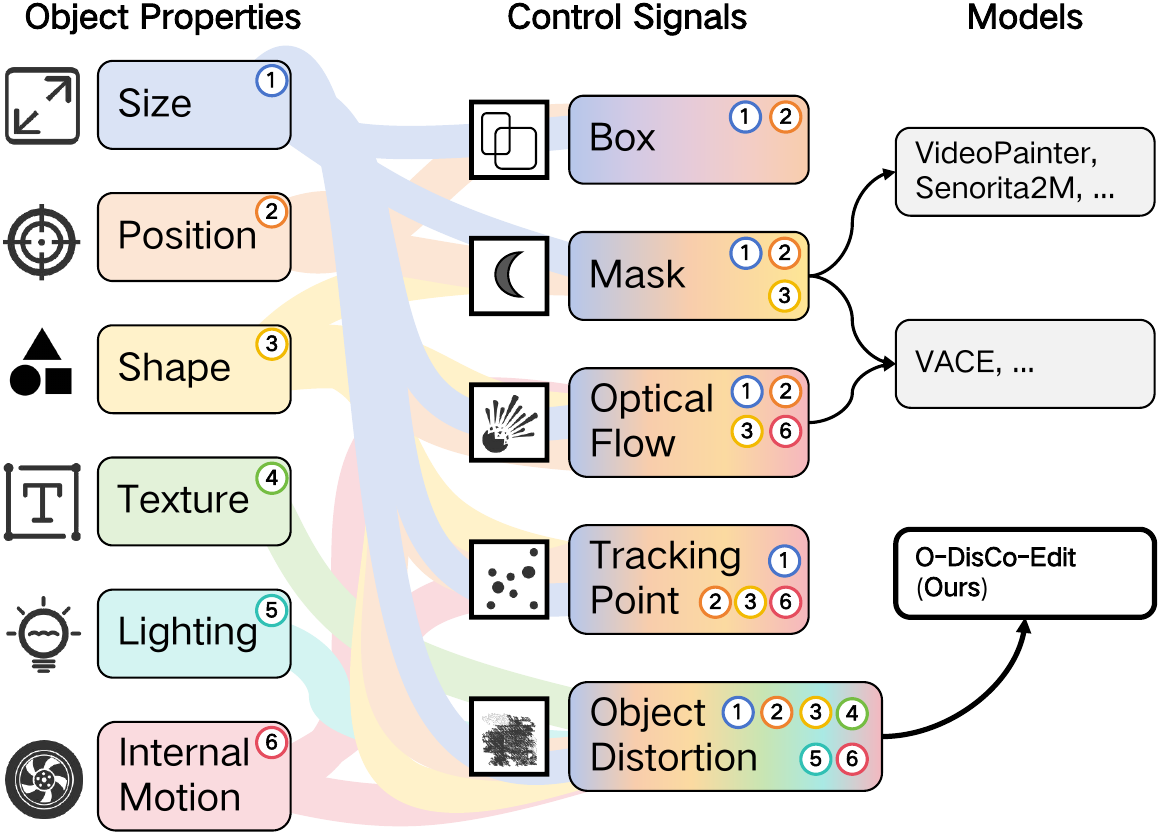}
    \caption{Comparisons of different object properties,  control signals, and models.}
    \label{intro-signal}
    \vspace{-0.2in}
\end{figure}

For controllable video editing,  single-task editing models incorporate additional control signals such as 2D bounding boxes~\citep{tu2025videoanydoor, yang2024direct, li2025magicmotion},  masks~\citep{zhao2025dreaminsert, yariv2025through}, optical flow~\citep{yin2023dragnuwa, wang2025consistent, liu2024stablev2v},  and tracking points~\citep{gu2025diffusion, wu2024safety, wang2025levitor} to improve control precision.  However, as shown in~\cref{intro-signal}, conditions like bounding boxes and masks provide limited information, thus  hindering fine-grained control for complex editing scenarios. 
Furthermore, video datasets with optical flow and tracking are scarce, and their extraction is often complex and prone to inaccuracies. These two issues make precise and intricate controllable editing difficult. 

Single-task editing models, as discussed above, are no longer sufficient to meet user diverse demands. 
Consequently, unified multi-task video editing approaches~\citep{liang2025omniv2v, li2025tokenmotion, zhang2025enabling, ye2025unic, jiang2025vace} have emerged, which can accomplish diverse editing tasks by introducing various signals. 
However, they typically demand complex training pipelines. 
This complexity necessitates the construction of specialized multi-task datasets and the design of task-specific modules (e.g., multiple DiT blocks), resulting in a large number of trainable parameters, as shown in~\cref{tab:traning_cogif}. 
Furthermore, it requires integrating various conditions across numerous training stages, often demanding tens of thousands of steps.

Despite their design incorporating diverse signals, most of multi-task video editing model are inflexible during inference, as they can generally process only one control condition at a time.  
This prevents the model from leveraging complementary cues from multiple signals, thereby hindering the flexible transition  between fine-grained and coarse-grained editing for the same task. 

To address these challenges, we propose a novel unified control signal:  the \textbf{o}bject \textbf{dis}tortion \textbf{co}ntrol (O-DisCo). 
This signal is generated by applying appropriate noise to the edited objects, effectively acting as a distortion signal for the reference video. 
As illustrated in~\cref{intro-signal}, all other control signals can similarly be viewed as specific types of reference video distortion. 
Therefore, by controlling the noise, O-DisCo inherently unifies all these signals into a single representation. 
For training, randomness is introduced into O-DisCo. 
This significantly simplifies training dataset construction and model design, saving substantial training resources, as shown in~\cref{tab:traning_cogif}. 
For inference, by adaptively manipulating the intensity and scope of O-DisCo's noise, our model can perform a wide range of tasks. 
While the above design primarily focuses on the edited regions, a ``copy-form'' preservation module is further designed to address the preservation of non-edited areas. 
Encapsulating these capabilities, we propose O-DisCo-Edit, a unified framework for versatile video editing. 

\begin{table}[!t]
\small
  \centering
  \caption{Comparison of training configurations for different models. * indicates that the majority of the module is used for training. ``Block'' refers  to a DiT block.}
\setlength{\tabcolsep}{0.6mm}
    \begin{tabular}{ccccc}
    \toprule
    Model & Dataset & Trainable Module & Steps & GPUs \\
    \midrule
    VACE  & Mutil-Task  & 8 Blocks & 200K  & 128 A100 \\
    Senorita & Mutil-Task  & 102 Blocks* & 4 epoch & \textbackslash{} \\
    VideoPainter & 390.3k & {2 Blocks, 1 LoRA} & 82K   & 64 V100 \\
    Ours  & 180k  & Two LoRAs & 7.55K & 8 A800 \\
    \bottomrule
    \end{tabular}%

\vspace{-1em}
\label{tab:traning_cogif}
\end{table}

Our comprehensive experiments confirm O-DisCo-Edit’s effectiveness and versatility across diverse tasks, including object removal, outpainting, and transfers of motion, lighting, color, and style. 
Specifically, O-DisCo-Edit consistently  surpasses the current state-of-the-art (SOTA) multi-task editing model, VACE~\citep{jiang2025vace}, on the majority of tasks. Notably, for the object removal task on the OmnimatteRF~\citep{lin2023omnimatterf} benchmark, our method also demonstrates superior performance compared to the specialized SOTA removal approach, MiniMax-Remover~\citep{zi2025minimax}.

Overall, the contribution of this work is summarized as follows:
\begin{itemize}
\item A novel unified control signal,  \textbf{o}bject \textbf{dis}tortion \textbf{co}ntrol (O-DisCo) is proposed to substantially reduce training resource demands and enable flexible, precise multi-task video editing from coarse to fine granularity.
\item We propose a ``copy-form'' preservation module for non-edited region preservation,  which enhances the model's ability to maintain unedited areas. 
\item Our proposed O-DisCo-Edit, achieving new SOTA performance across diverse tasks, offers a novel perspective for developing unified video editing frameworks.
\end{itemize}

\section{Related Work}
\label{sec:related}                             
\begin{figure*}[t]
    \centering
    \includegraphics[width=1.0\linewidth]{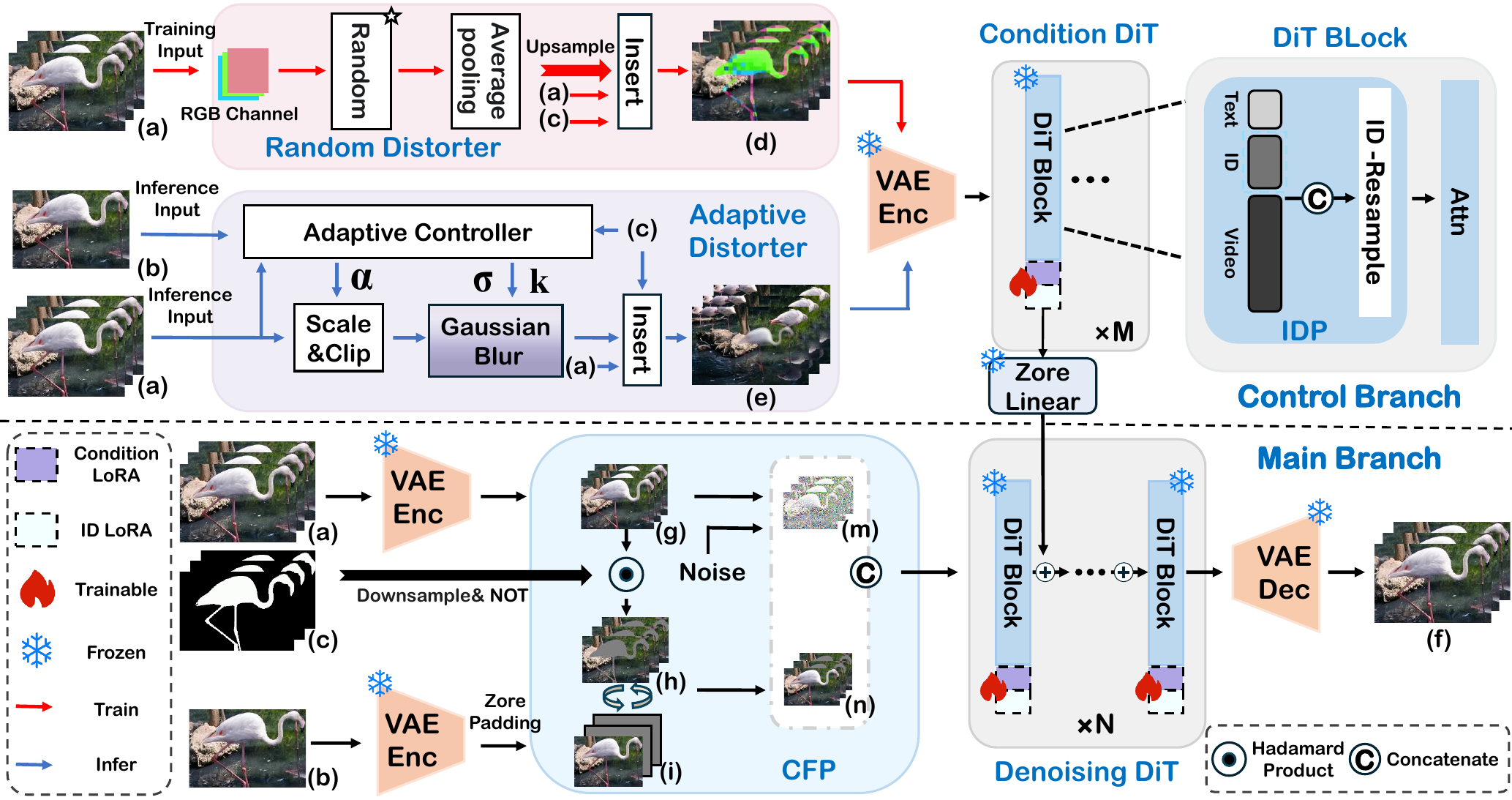}
    \caption{The framework of the proposed O-DisCo-Edit. (a) Reference video. (b) Reference image (first frame during training, edited image during inference). (c) Masks. (d) R-O-DisCo. (e) A-O-DisCo. (f) Generated video. (g) Latent of  reference video. (h) Latent of the preserved region. (i) Image latent with zero-padding. (m) Noisy Latent. (n) Image Latent with the latent of preserved region. $\alpha$ represents the contrast, $\sigma$ represents the intensity of the added noise, and ${k}$ is the size of the gaussian blur kernel. The adaptive distorter generates A-O-DisCo for inference, and the random distorter generates R-O-DisCo for training. The CFP ensures the preservation of unedited areas. The IDP maintains object appearance consistency.}
    \label{fig:framework}
    \vspace{-0.1in}
\end{figure*}
\noindent{\textbf{Single-Task Video Editing and Control Signals.}}~Video editing tasks frequently require additional control signals (e.g.,  masks, poses, optical flows, tracking points) to modify reference video attributes. 
 VideoAnydoor~\citep{tu2025videoanydoor} introduces masks and tracking points for object insertion, while DiffuEraser~\citep{gu2025diffusion} leverage masks for object removal. 
Follow-your-Canvas~\citep{chen2024follow} employs 2D bounding boxes for outpainting, and ReCamMaster~\citep{bai2025recammaster} uses camera trajectories for camera control.

\vspace{0.5em}
\noindent{\textbf{{Multi-Task Video Editing.}}~Growing demands for creative versatility have driven the development of multi-task video editing. 
VACE~\citep{jiang2025vace} integrates sophisticated signals like optical flow and masks with a context embedder and adapter to perform tasks such as swap, animation, and outpainting. 
Similarly, Senorita~\citep{zi2025se} utilizes masks, canny edges, and other cues, coupled with four specialized expert models, to achieve tasks like addition, removal, and swap. 
Therefore, multi-task editing often demands complex training pipelines with diverse signals, specialized modules, and multi-stage training~\citep{ye2025unic, liang2025omniv2v, jiang2025vace, zi2025se}. 
In contrast, our proposed unified O-DisCo signal enables multi-task completion with significantly fewer training resources.

\vspace{0.5em}
\noindent{\textbf{Adaptive Inference.}}~Current video editing models~\citep{jiang2025vace, zi2025se, liang2025omniv2v, ye2025unic, tu2025videoanydoor} typically rely on a single control signal during inference, which limits their adaptability for multi-grained editing. 
Adaptive inference, conversely, allows models to adjust outputs dynamically based on varying reference videos, images, or prompts. 
This is common in training-free models; for example, LMP~\citep{chen2025lmp} applies attention values as a loss to optimize hidden states for appearance similarity. 
Likewise, the  authors in~\citep{zhang2025training} defines motion consistency loss for gradient descent on noisy latent vectors to achieve motion transfer. 
Building on this principle, O-DisCo-Edit also leverages adaptive inference through the proposed O-DisCo, which dynamically adjusts injected noise based on reference videos and images for multi-grained control. 

\section{Method}
\label{sec:method}
As shown in~\cref{fig:framework}, our approach introduces a first-frame-guided video editing model, building upon the CogVideoX-I2V architecture~\citep{yang2024cogvideox}. 
The object distortion control (O-DisCo), derived via the distorter, is fed into the VAE and conditional DiT for precise control over edited regions.  
Concurrently, a ``copy-form'' preservation (CFP) module processes the reference image and video, which then provides their latent output to the denoising DiT for  preservation of non-edited areas. 
Additionally, an identity preservation (IDP) module is proposed to enhance ID fidelity within edited regions.  
Subsequent sections detail O-DisCo's construction and the design of the CFP and IDP modules.
\subsection{Random Object Distortion Control}
During the training phase, we apply a random distoter to generate random object distortion control (R-O-DisCo).  
As shown in the top-left part of~\cref{fig:framework}, we intentionally distort the colors of the reference video $ V_\text{ref}\in \mathbb R^{F \times H \times W \times 3}$ to prevent the model from simply replicating original color information, where $F$ is the frame number, $H$ and $W$ are the height and width of the reference video.  
This involves applying \textbf{random} arithmetic operations to each RGB channel, and the resulted color-distorted video $V_\text{cd}\in \mathbb R^{F \times H \times W \times 3}$ is formulated as:
\begin{equation} 
\label{eq:color_distorter}
\begin{array}{c}
    V_\text{cd}[:, :, :, i] = \text{clip}(V_\text{ref}[:, :, :, i] \star a_{i}),  \ i\in \{0,  1,  2\},\\
\end{array}
\end{equation}
where $a_{i}, i\in \{0, 1, 2\}$ are random real numbers, with their specific ranges detailed in the Appendix A. 
Moreover, $\star$ denotes a randomly chosen arithmetic operation (addition, subtraction, multiplication, or division). 
The clip operation constrains output values to the range $[0, 255]$. 

After that, $V_{\text{cd}}$ is downsampled via average pooling and then upsampled using nearest-neighbor interpolation, both by a factor of $L \in \mathbb{Z}$. 
$L$ is a randomly sampled integer resampling factor (more details in Appendix A). 
This operation intentionally disrupts fine-grained structural details, producing a mosaic-like effect video $V_{\text{cdm}}\in \mathbb R^{F \times H \times W \times 3}$. 
Consequently, the model is compelled to primarily learn video generation guided by the first frame's appearance, rather than relying on the precise visual information of $V_\text{cd}$.

Finally, We utilize masks $M \in \mathbb {R}^{F  \times H \times W \times 1}$ to insert the edited object from $V_\text{cdm}$ into $V_\text{ref}$, which produces the R-O-DisCo $V_\text{RODC}$ ((d) in~\cref{fig:framework}) during the training stage:
\begin{equation}
\label{insert}
\begin{array}{c}
   V_\text{RODC} = V_\text{cdm}\odot M +V_\text{ref}\odot (\mathbf{1}-M ), 
\end{array}
\end{equation} 
where $\odot$ represents the Hadamard product.  

Overall,  during the training phase,  we enhance the model's robustness and task adaptability by increasing the randomness of O-DisCo. 

\subsection{Adaptive Object Distortion Control} 
During inference phase,  our model adapts to specific tasks or instructions via adaptive object distortion control (A-O-DisCo) (highlighted by (e) in~\cref{fig:framework}), which is implemented by an adaptive distorter.   
It is achieved through contrast modification (scaling and clipping) and dynamic noise injection within the  editable regions. The process is formally represented by the following equations:
\begin{equation}
\begin{aligned}
    &V_{\text{c}}(f, x, y) = \text{clip}\big(\alpha \cdot V_\text{ref}(f, x, y)\big),  \\
    &V_{\text{cn}}(f, x, y) = \sum_{i=-b}^{b} \sum_{j=-b}^{b} V_\text{c}(f, x+i,  y+j) \cdot G_{\text{norm}}(i, j;\sigma),  \\
    &V_{\text{AODC}} =   V_{\mathrm{cn}} \odot M + V_{\text{ref}} \odot (\mathbf{1}-M), 
\end{aligned}\label{AODC}
\end{equation}
where $V_{\text{ref}}(f, x, y)$ denotes the pixel value at coordinates $(x, y)$ in the $f$-th frame of the reference video. 
$V_\text{c}(x, y)$ and $V_\text{cn}(x, y)$ represent the  $V_{\text{ref}}$ after scale\&clip  and gaussian blur, respectively.   $G_{\text{norm}}(i, j;\sigma)$ is the normalized gaussian blur kernel. $\alpha$ represents the contrast, $\sigma$ is the noise intensity, and $k=2b+1$ is the gaussian kernel size. 

The adaptive controller determines suitable values for $\alpha$, $\sigma$,  and $k$ by calculating two similarities: $\mathbf{Sim}_{i}$,  the edited region's edge map similarity between the reference image and the reference video's first frame; $\mathbf{Sim}_\text{v}$, the intra-frame similarity within the reference video's edited region edge map. Empirically, fitting these three parameters using a quadratic polynomial of two similarity  yields superior results (see specific formulas in Appendix A). Finally,  the A-O-DisCo $V_\text{AODC}$ obtained for each task is shown in the last line of~\cref{tab:Task_condition}.   Specifically,  during inference for  object removal and outpainting (R\&O),  we set $V_\text{AODC}$ to zero,  ensuring that no additional information is introduced.   This allows the model to generate the video  based on other condition,  thereby reducing artifacts.  

\subsection{``Copy-Form''  Preservation Module}

Many video editing methods~\citep{jiang2025vace, bian2025videopainter, liang2025omniv2v} typically integrate non-edited regions with the control signals via extra branch. However, such integration often leads to mutual interference between the preserved regions and control signals, thereby limiting editing flexibility.
Instead,  we propose the ``copy-form''  preservation (CFP) module  illustrated~\cref{fig:framework}, which enhances editing flexibility by integrating the non-edited regions directly into the main branch of the network. Detailedly, CFP  replaces conventional zero-padding (denoted as (i) in~\cref{fig:framework}) with the latent  of the preserved region $z^{\text{v}'}_\text{p}$ (marked as (h) in~\cref{fig:framework}), to obtain $z_\text{images}$ (denoted as (n) in~\cref{fig:framework}). This process is expressed as:
\begin{equation}
    \begin{aligned}
\label{copy_form}
    & z^{\text{v}'}_\text{p}=z^\text{v}_\text{ref} \odot (\mathbf{1}-z_\text{mask}[1:]), \\
    & z_\text{images} =[z^\text{i}_\text{ref},  z^{\text{v}'}_\text{p}], 
\end{aligned} 
\end{equation}
where $z^\text{v}_\text{ref}$ denotes the latent of the reference video, $z_\text{mask}$ represents the downsampled binary mask (with the same shape as $z^\text{v}_\text{ref}$), $z^\text{i}_\text{ref}$ is reference image latent. $[1:]$ corresponds to a slicing operation,  and $[, ]$ signifies tensor concatenation.  The $z_\text{images}$ for each task are shown in~\cref{tab:Task_condition}.  Notably, the CFP module achieves preservation of non-edited regions with an effect similar to ``first-frame copying''.

\begin{table}[!t]
  \centering
  \setlength{\tabcolsep}{1mm} 
    \caption{Adaptive inference conditions for different tasks. R\&O in the first line means object removal and outpainting.}
  \begin{tabular}{l c c c } 
    \toprule
    \makecell{{Condition}} 
    & \makecell{{R\&O}}
    & \makecell{{Style Tranfer}}
    & \makecell{{Other Tasks}}
    \\
    \midrule
    $z_\text{images}$
    & $[z_{\text{ref}}^\text{i}, z_{\text{p}}^{\text{v}'}]$
    & $[z_{\text{ref}}^\text{i}, \mathbf{0}]$
    & $[z_{\text{ref}}^\text{i}, z_{\text{p}}^{\text{v}'}]$
    \\
    \midrule
    \multirow{2}{*}{$V_\text{AODC}$}
    & \multirow{2}{*}{$\mathbf{0}$}
    & \multirow{2}{*}{\makecell{$V_\text{cn}\odot M$ \\ $+V_\text{ref}(1-M)$  }}
    & \multirow{2}{*}{\makecell{$V_\text{cn}\odot M$ \\ $+V_\text{ref}(1-M)$  }}\\
    & & \multicolumn{1}{c}{} &  \multicolumn{1}{c}{} \\[-\normalbaselineskip]\\
  
    \bottomrule
  \end{tabular}

  \vspace{-1em}
  \label{tab:Task_condition}
\end{table}

\subsection{Identity Preservation Module}
To mitigate object appearance changes during complex motion or occlusion, we design the identity preservation (IDP) module, illustrated in the upper right corner of~\cref{fig:framework}. Specifically, we extract position-agnostic tokens (ID tokens) from the reference image's edited regions and concatenate them with text tokens. Akin  to text tokens,  ID tokens act as a global guide which make the model leverage  ID information throughout the video generation. Further enhancing the model's focus on ID consistency, we introduce ID-Resample to extract the key (K) and value (V) vectors from the edited regions of the generated video. These are concatenated with the K and V vectors of the original generated video, and the process compels the model to reinforce ID consistency within the edited regions.

\begin{table*}[htbp]
  \centering
 \caption{Comparison of different models on various tasks using our benchmark (OmnimatteRF benchmark for object removal). The evaluation includes automatic scoring and a manual user study. The best results are  in \textbf{bold}, while the second best are \underline{underlined}. ``Preservation'' means  non-edited region preservation, ``TC'' denotes temporal consistency, ``EC'' represents editing completeness, and ``VQ'' stands for visual quality.}
 \small
 \setlength{\tabcolsep}{1mm}
  \centering
    \begin{tabular}{c|c|ccccccc|cc}
    \toprule
    Metrics &  & \multicolumn{4}{c}{Video Quality} & \multicolumn{2}{c}{Removal Capability} & Normalized    & \multicolumn{2}{c}{User Stuty} \\
    \midrule
    Task  & Method & TC$\uparrow$     & FVD$\downarrow$   & PSNR$\uparrow$   & SSIM$\uparrow$   & SSIM$_\text{E}\uparrow$  & PSNR$_\text{E}\uparrow$  & Avg. Score$\uparrow$  & EC$\uparrow$     & VQ$\uparrow$  \\
    \midrule
    \multirow{4}[1]{*}{\makecell{ (a.1) \\Object\\ Removal (49)}} & DiffuEraser & 0.9964 & 422.7 & 27.89 & \underline{0.9207} & 0.9713 & 34.09 & 0.2682 & 3.122 & 2.867 \\
          & MiniMax & \underline{0.9973} & \underline{373.2} & 27.15 & 0.8732 & \textbf{0.9737} & \underline{34.87} & 0.4816 & \underline{3.567} & \underline{3.333} \\
          & propainter & 0.9971 & {410.3} & \textbf{28.30} & \textbf{0.9224} & 0.9715 & 34.12 & \underline{0.4844} & 3.044 & 2.822 \\
          & \cellcolor[rgb]{ .816,  .816,  .816}O-DisCo-Edit & \cellcolor[rgb]{ .816,  .816,  .816}\textbf{0.9974} & \cellcolor[rgb]{ .816,  .816,  .816}\textbf{300.3} & \cellcolor[rgb]{ .816,  .816,  .816}\underline{28.05} & \cellcolor[rgb]{ .816,  .816,  .816}0.8751 & \cellcolor[rgb]{ .816,  .816,  .816}\underline{0.9730} & \cellcolor[rgb]{ .816,  .816,  .816}\textbf{35.43} & \cellcolor[rgb]{ .816,  .816,  .816}\textbf{0.7553} & \cellcolor[rgb]{ .816,  .816,  .816}\textbf{3.967} & \cellcolor[rgb]{ .816,  .816,  .816}\textbf{3.689} \\
    \midrule
    \multirow{5}[1]{*}{\makecell{ (a.2) \\Object \\Removal (33)}} & VACE 1.3B & 0.9934 & 1376  & 23.46 & \underline{0.8508} & 0.9551 & 26.17 & 0.4233 & 2.011 & 1.911 \\
          & VACE 14B & 0.9896 & 2085  & 22.29 & 0.8372 & 0.9439 & 24.28 & 0.1316 & 1.578 & 1.444 \\
          & Senorita & \underline{0.9962} & \underline{662.2} & \underline{26.24} & 0.8387 & \underline{0.9681} & \underline{32.77} & \underline{0.7058} & \underline{3.311} & \underline{3.156} \\
          & VideoPainter & 0.9871 & 2403  & 21.28 & 0.8303 & 0.9452 & 23.42 & 0.0072 & 1.744 & 1.578 \\
          & \cellcolor[rgb]{ .816,  .816,  .816}O-DisCo-Edit & \cellcolor[rgb]{ .816,  .816,  .816}\textbf{0.9969} & \cellcolor[rgb]{ .816,  .816,  .816}\textbf{360.1} & \cellcolor[rgb]{ .816,  .816,  .816}\textbf{28.28} & \cellcolor[rgb]{ .816,  .816,  .816}\textbf{0.8719} & \cellcolor[rgb]{ .816,  .816,  .816}\textbf{0.9740} & \cellcolor[rgb]{ .816,  .816,  .816}\textbf{36.05} & \cellcolor[rgb]{ .816,  .816,  .816}\textbf{1.000} & \cellcolor[rgb]{ .816,  .816,  .816}\textbf{3.956} & \cellcolor[rgb]{ .816,  .816,  .816}\textbf{3.689} \\
    \midrule \midrule 
    Metrics &       & \multicolumn{3}{c}{Video Quality} & \multicolumn{1}{c}{Alignment} & \multicolumn{2}{c}{Preservation} & Normalized    & \multicolumn{2}{c}{User Stuty} \\
    \midrule
    
    Task  & Method & TC$\uparrow$    & FVD$\downarrow$     & PSNR$\uparrow$   & CLIP-T$\uparrow$  & PSNR$_\text{P}\uparrow$ & SSIM$_\text{P}\uparrow$ & Avg. Score$\uparrow$  & EC$\uparrow$     & VQ$\uparrow$  \\
    \midrule
    
    \multirow{5}[2]{*}{ \makecell{(b) \\ Outpainting}} & VACE 1.3B & 0.9976 & \underline{88.03}  & 25.11 & 11.92 & 31.62 & 0.9383 & \underline{0.6801} & \underline{4.244} & \underline{3.933} \\
          & VACE 14B & \underline{0.9977} & 88.75 & 23.52 & 12.03 & 30.08 & 0.9381 & 0.4972 & 4.178 & 3.911 \\
          & Senorita & \textbf{0.9978} & 177.1 & \underline{25.20} & \textbackslash{} & 30.90  & 0.9262 & 0.4784 & 2.889 & 2.600 \\
          & VideoPainter & 0.9958 & 325.4 & 24.54 & \textbf{12.95} & \underline{31.81} & \underline{0.9442} & 0.3384 & 2.156 & 1.978 \\
          & \cellcolor[rgb]{ .816,  .816,  .816}O-DisCo-Edit & \cellcolor[rgb]{ .816,  .816,  .816}\textbf{0.9978} & \cellcolor[rgb]{ .816,  .816,  .816}\textbf{77.03} & \cellcolor[rgb]{ .816,  .816,  .816}\textbf{26.43} & \cellcolor[rgb]{ .816,  .816,  .816}\underline{12.18} & \cellcolor[rgb]{ .816,  .816,  .816}\textbf{33.87} & \cellcolor[rgb]{ .816,  .816,  .816}\textbf{0.9466} & \cellcolor[rgb]{ .816,  .816,  .816}\textbf{1.000} & \cellcolor[rgb]{ .816,  .816,  .816}\textbf{4.289} & \cellcolor[rgb]{ .816,  .816,  .816}\textbf{4.067} \\
        \midrule \midrule 
    Metrics &       & \multicolumn{2}{c}{Video Quality} & \multicolumn{2}{c}{Alignment} & \multicolumn{2}{c}{Preservation} & Normalized    & \multicolumn{2}{c}{User Stuty} \\
    \midrule
    Task  & Method & TC$\uparrow$     & ArtFID$\downarrow$ & CLIP-T$\uparrow$  & CLIP-I$_\text{E}\uparrow$  & PSNR$_\text{P}\uparrow$  & SSIM$_\text{P}\uparrow$  & Avg. Score$\uparrow$  & EC$\uparrow$     & VQ$\uparrow$  \\
    \midrule
    
    \multicolumn{1}{c|}{\multirow{5}[2]{*}{\makecell{ (c) \\ Object Internal  \\ Motion Transfer}}} & VACE 1.3B & \textbf{0.9946} & 7.329 & 18.75 & 93.73 & \textbf{36.41} & \textbf{0.9586} & \underline{0.8515} & 3.011 & 2.533 \\
          & VACE 14B & 0.9937 & 7.025 & \underline{18.96} & 93.39 & 34.47 & \underline{0.9582} & 0.7641 & \underline{3.122} & \underline{2.800} \\
          & Senorita & \underline{0.9940} & \textbf{6.628} & \textbackslash{} & \underline{93.94} & 29.45 & 0.8477 & 0.5229 & 3.022 & 2.333 \\
          & VideoPainter & 0.9908 & 8.201 & \textbf{19.47} & 91.57 & \underline{36.32} & 0.9410 & 0.3657 & 2.300   & 1.822 \\
          & \cellcolor[rgb]{ .816,  .816,  .816}O-DisCo-Edit & \cellcolor[rgb]{ .816,  .816,  .816}0.9927 & \cellcolor[rgb]{ .816,  .816,  .816}\underline{6.712} & \cellcolor[rgb]{ .816,  .816,  .816}18.82 & \cellcolor[rgb]{ .816,  .816,  .816}\textbf{94.64} & \cellcolor[rgb]{ .816,  .816,  .816}35.88 & \cellcolor[rgb]{ .816,  .816,  .816}0.9530 & \cellcolor[rgb]{ .816,  .816,  .816}\textbf{0.8639} & \cellcolor[rgb]{ .816,  .816,  .816}\textbf{4.178} & \cellcolor[rgb]{ .816,  .816,  .816}\textbf{3.756} \\
    \midrule 
    \multicolumn{1}{c|}{\multirow{5}[2]{*}{\makecell{ (d) \\Lighting  \\ Transfer}}} & VACE 1.3B & \textbf{0.9964} & \textbf{5.991} & 20.26 & 95.30  & 31.46 & \underline{0.9292} & \underline{0.7700} & 3.067 & 2.644 \\
          & VACE 14B & \underline{0.9958} & 6.187 & 20.35 & 95.32 & 30.69 & 0.9261 & 0.5489 & \underline{3.411} & \underline{3.067} \\
          & Senorita & \textbf{0.9964} & 6.478 & \textbackslash{} & \textbf{96.15} & 28.71 & 0.9011 & 0.4000   & 3.033 & 2.400 \\
          & VideoPainter & 0.9951 & 6.378 & \textbf{21.92} & 94.76 & \underline{32.93} & \textbf{0.9325} & 0.4160 & 2.911 & 2.489 \\
          & \cellcolor[rgb]{ .816,  .816,  .816}O-DisCo-Edit & \cellcolor[rgb]{ .816,  .816,  .816}0.9956 & \cellcolor[rgb]{ .816,  .816,  .816}\underline{6.043} & \cellcolor[rgb]{ .816,  .816,  .816}\underline{20.86} & \cellcolor[rgb]{ .816,  .816,  .816}\underline{96.05} & \cellcolor[rgb]{ .816,  .816,  .816}\textbf{33.54} & \cellcolor[rgb]{ .816,  .816,  .816}0.9285 & \cellcolor[rgb]{ .816,  .816,  .816}\textbf{0.8157} & \cellcolor[rgb]{ .816,  .816,  .816}\textbf{3.978} & \cellcolor[rgb]{ .816,  .816,  .816}\textbf{3.689} \\
     \midrule 
    \multirow{5}[2]{*}{\makecell{ (e) \\Change Color}} & VACE 1.3B & \underline{0.9955} & 8.150  & 11.89 & 97.16 & 30.55 & \underline{0.9056} & \underline{0.7838} & 3.633 & 3.244 \\
          & VACE 14B & 0.9954 & 8.485 & 11.63 & 96.55 & 29.76 & 0.9036 & 0.5703 & 3.456 & 3.089 \\
          & Senorita & \textbf{0.9959} & \textbf{8.002} & \textbackslash{} & \textbf{97.67} & 27.52 & 0.8724 & 0.6000   & \textbf{4.033} & \textbf{3.711} \\
          & VideoPainter & 0.9943 & 9.388 & \textbf{12.89} & 96.41 & \underline{30.99} & \textbf{0.9136} & 0.3996 & 3.633 & 3.267 \\
          & \cellcolor[rgb]{ .816,  .816,  .816}O-DisCo-Edit & \cellcolor[rgb]{ .816,  .816,  .816}\underline{0.9955} & \cellcolor[rgb]{ .816,  .816,  .816}\underline{8.008} & \cellcolor[rgb]{ .816,  .816,  .816}\underline{11.94} & \cellcolor[rgb]{ .816,  .816,  .816}\underline{97.49} & \cellcolor[rgb]{ .816,  .816,  .816}\textbf{31.00} & \cellcolor[rgb]{ .816,  .816,  .816}0.9049 & \cellcolor[rgb]{ .816,  .816,  .816}\textbf{0.8787} & \cellcolor[rgb]{ .816,  .816,  .816}\underline{3.944} & \cellcolor[rgb]{ .816,  .816,  .816}\underline{3.689} \\
    \midrule 
    Task  & Method & TC$\uparrow$     & FVD$\downarrow$   & CLIP-T$\uparrow$  & CLIP-I$_\text{E}\uparrow$  & PSNR$_\text{P}\uparrow$  & SSIM$_\text{P}\uparrow$  & Avg. Score$\uparrow$  & EC$\uparrow$     & VQ$\uparrow$  \\
    \midrule
    \multirow{5}[2]{*}{\makecell{ (f) \\Swap}} & VACE 1.3B & \underline{0.9843} & \underline{688.2} & 14.73 & 91.28 & \underline{26.36} & 0.8062 & \textbf{0.7068} & 3.467 & 2.956 \\
          & VACE 14B & 0.9841 & \textbf{642.3} & 14.83 & 91.03 & 25.76 & 0.8041 & \underline{0.6959} & \underline{3.556} & \underline{3.089} \\
          & Senorita & \textbf{0.9845} & 803.7 & \textbackslash{} & \textbf{92.76} & 23.66 & 0.7436 & 0.4000   & 3.456 & 2.956 \\
          & VideoPainter & 0.9815 & 731.2 & \textbf{15.91} & 90.05 & \textbf{27.51} & \textbf{0.8295} & 0.4899 & 2.967 & 2.178 \\
          & \cellcolor[rgb]{ .816,  .816,  .816}O-DisCo-Edit  &  \cellcolor[rgb]{ .816,  .816,  .816}0.9839 & \cellcolor[rgb]{ .816,  .816,  .816}711.8 & \cellcolor[rgb]{ .816,  .816,  .816}\underline{15.27} & \cellcolor[rgb]{ .816,  .816,  .816}\underline{91.84} & \cellcolor[rgb]{ .816,  .816,  .816}26.25 & \cellcolor[rgb]{ .816,  .816,  .816}\underline{0.8098} & \cellcolor[rgb]{ .816,  .816,  .816}0.6950 & \cellcolor[rgb]{ .816,  .816,  .816}\textbf{4.033} & \cellcolor[rgb]{ .816,  .816,  .816}\textbf{3.689} \\
    \midrule 
    \multirow{5}[2]{*}{\makecell{ (g) \\Addition}} & VACE 1.3B & 0.9862 & 512.7 & \underline{21.04} & 92.92 & 27.47 & \underline{0.8094} & 0.3419 & 3.370  & 2.489 \\
          & VACE 14B & \underline{0.9873} & \underline{398.9} & \textbf{21.41} & 92.58 & 26.86 & 0.8082 & 0.3621 & 3.200   & 2.578 \\
          & Senorita & \textbf{0.9891} & \textbf{316.8} & \textbackslash{} & \textbf{95.39} & 27.89 & 0.7934 & \textbf{0.7375} & \underline{3.496} & \underline{3.089} \\
          & VideoPainter & 0.9836 & 560.9 & 20.77 & 93.64 & \textbf{28.36} & \textbf{0.8246} & 0.4754 & 3.104 & 2.911 \\
          & \cellcolor[rgb]{ .816,  .816,  .816}O-DisCo-Edit & \cellcolor[rgb]{ .816,  .816,  .816}0.9871 & \cellcolor[rgb]{ .816,  .816,  .816}448.3 & \cellcolor[rgb]{ .816,  .816,  .816}21.03 & \cellcolor[rgb]{ .816,  .816,  .816}\underline{95.37} & \cellcolor[rgb]{ .816,  .816,  .816}\underline{28.03} & \cellcolor[rgb]{ .816,  .816,  .816}0.8048 & \cellcolor[rgb]{ .816,  .816,  .816}\underline{0.6470} & \cellcolor[rgb]{ .816,  .816,  .816}\textbf{4.037} & \cellcolor[rgb]{ .816,  .816,  .816}\textbf{3.822} \\
    \bottomrule
    \end{tabular}%
\label{tab:comparison}
\end{table*}

\section{Experiment}

\noindent\textbf{Implementation Details.}~Training dataset: we utilize approximately $180$k video-mask pairs from the Senorita-2M grounding dataset~\citep{zi2025se}. 
All video-mask pairs are center-cropped and resized to a $720 \times 480$ resolution with a length of $49$ frames. Moreover, prompts for masked regions are generated via Qwen2.5-VL-7B~\citep{bai2023qwen}. 
Two-stage training: Our model builds on the frozen pre-trained weights of Diffusion as Shader~\citep{gu2025diffusion}. 
Firstly, condition LoRA is trained with the random distorter and CFP module ($2400$ steps); 
Secondly, the IDP module's dedicated ID LoRA is trained ($5150$ steps). 
All stages employ AdamW optimization (learning rate $1 \times 10^{-4}$) on $8$ A800 GPUs with gradient accumulation for a batch size of $32$.  

\vspace{0.5em}
\noindent\textbf{Baseline Methods.}~For the majority of tasks, we select the SOTA unified video editing methods VACE~\citep{jiang2025vace},  VideoPainter~\citep{bian2025videopainter}, and Senorita ~\citep{zi2025se} as our primary baselines.  
Additionally, for the object removal, we include DiffuEraser~\citep{li2025diffueraser}, MiniMax-Remover (MiniMax)~\citep{zi2025minimax}, and Propainter~\citep{zhou2023propainter} as extra baselines.  
For the style transfer, Senorita~\citep{zi2025se} is chosen. 

\vspace{0.5em}
\noindent\textbf{Benchmarks.}~We curated a benchmark  from DIVAS~\citep{pont20172017} and VPData~\citep{bian2025videopainter}, specifically targeting challenging scenarios with internal motion, lighting variations, and complex object movements. 
For prompts, Senorita utilized the dedicated prompt, as required by its inference process, while others used same prompts from Qwen2.5VL-7B~\citep{bai2023qwen}. 
Additionally, the reference image was the first frame edited using either HiDream-E1~\citep{cai2025hidream} or commercial models~\footnote{\url{https://jimeng.jianying.com/}\label{fn:label1}}. 
Subsequently, the edited frame  served as input for O-DisCo-Edit and multi-task baselines.  
For fair comparison with Senorita, only the first $33$ frames of generated videos are evaluated. 
In parallel, OmnimatteRF~\citep{lin2023omnimatterf} is selected as a benchmark for the object removal task. 
Further details about our benchmark's construction are provided in Appendix B.

\vspace{0.5em}
\noindent\textbf{Metrics.}~The evaluation includes automatic scoring and a manual user study. 
Automatic scoring metrics:
(1) Non-Edited Region Preservation: Fidelity in unedited regions is assessed using  PSNR ($\text{PSNR}_\text{P}$) and SSIM ($\text{SSIM}_\text{P}$).
(2) Alignment: CLIP Similarity~\citep{wu2021godiva} (CLIP-T) measures semantic consistency between the generated video and its caption. Appearance consistency~\citep{zhang2025flexiact} ($\text{CLIP-I}_\text{E}$) between the output video and the reference image is calculated within the edited regions.
(3)  Video Generation Quality: Overall video quality is assessed via FVD~\citep{unterthiner2018towards}, ArtFID~\citep{wright2022artfid}, PSNR, SSIM, and temporal consistency (TC)~\citep{zhang2025flexiact, chen2023control}.
(4)  Normalized Average Score: This score is obtained by following the work of~\cite{huang2024vbench} using Min-Max Normalization, with all metrics (excluding CLIP-T) weighted equally.
Specifically, for the style transfer task, CFSD~\citep{chung2024style} is applied to evaluate the preservation of the reference video's content. Meanwhile, for the object removal task, we measure   removal ability by calculating the SSIM (SSIM$_\text{E}$) and PSNR (PSNR$_\text{E}$) between the edited regions of the output video and the corresponding background video.
Manual Assessment: The mean opinion score (MOS) is adopted, focusing on editing completeness (EC) and video quality (VQ). 
Anonymized generated data is randomly distributed to participants for $1$-$5$ scale scoring. More details are in the Appendix B.

\begin{figure*}[htbp]
    \centering
    \includegraphics[width=1.0\linewidth]{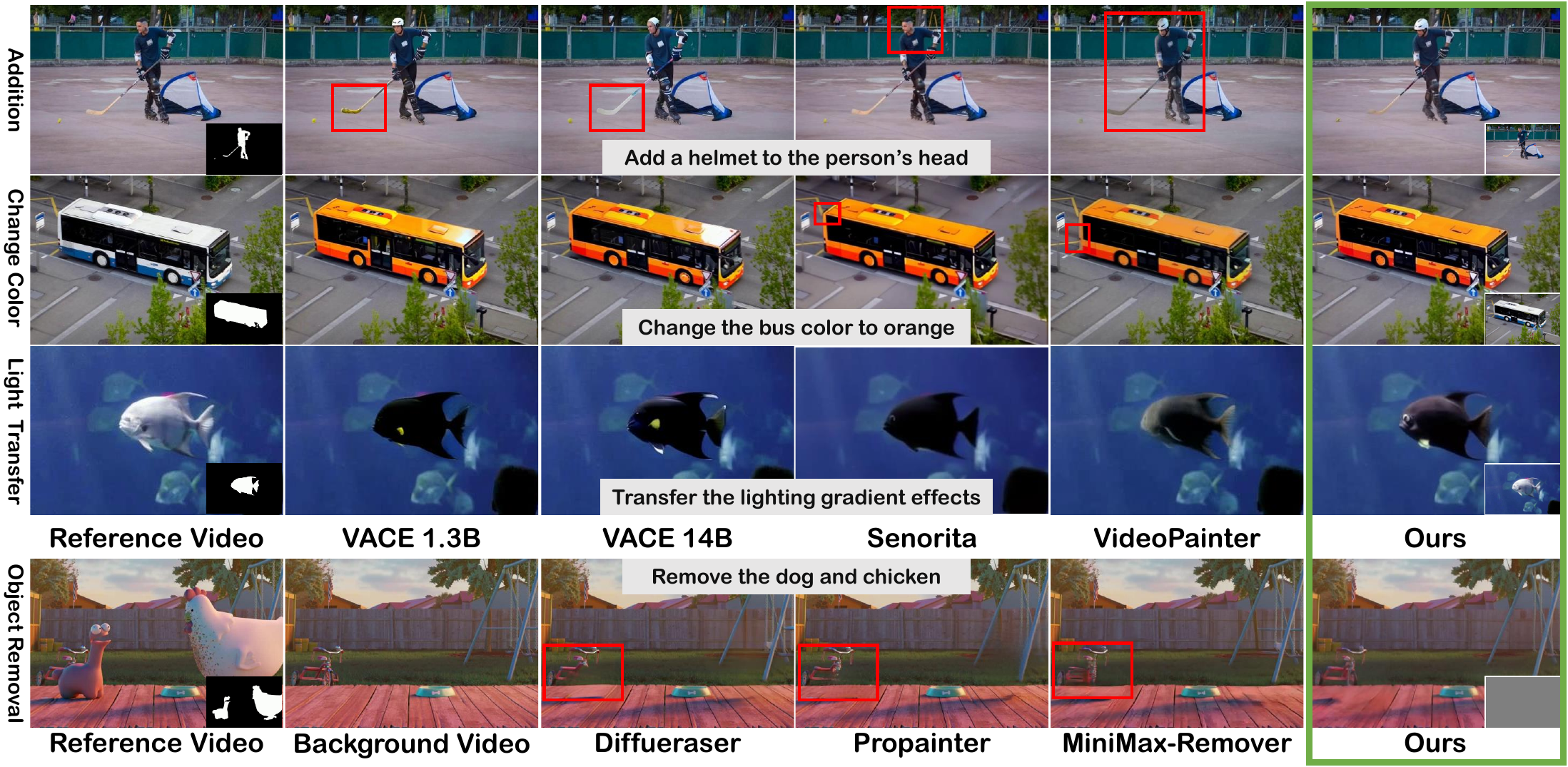}
    \caption{Our O-DisCo-Edit method is compared against other baselines for addition, color change, light transfer, and object removal. The bottom right of the reference video displays the input masks utilized by all models, while the corresponding position in our results highlights the A-O-DisCo required by our approach.
}
    \label{fig:Add_remove_change_color}
\end{figure*}


\subsection{Comparison with State-of-the-Arts}
As shown in~\cref{tab:comparison}, we conduct comprehensive comparisons between O-DisCo-Edit and baselines  across multiple tasks including  object removal, outpainting, object internal motion transfer, lighting transfer, color change, swap, addition, and style transfer. Our method demonstrates superior performance across all these tasks.

\vspace{0.5em}
\noindent\textbf{Object Removal.}~Quantitatively, our method obtains  optimal results in both the Remove ($49$) ($49$-frame videos) and Remove ($33$) ($33$-frame videos) settings (\cref{tab:comparison} (a)). As shown in~\cref{fig:Add_remove_change_color}, O-DisCo-Edit successfully avoids the background damage seen in Propainter and DiffuEraser, as well as the bicycle overlap present in MiniMax-Remover. 
In comparison with multi-task models in~\cref{fig:remove_style_inside_swap}, baselines consistently exhibit prominent artifacts, which indicate unsuccessful removal.

\begin{figure*}[htbp]
    \centering
    \includegraphics[width=1.0\linewidth]{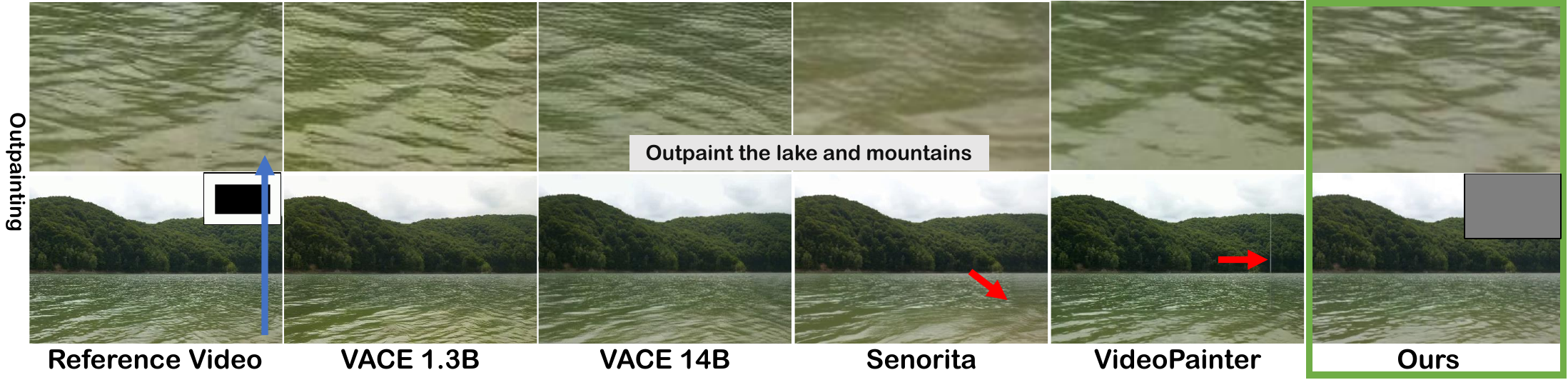}
    \caption{A comparison of O-DisCo-Edit and other baselines on the outpainting task. In the second row, the top right of the reference video displays the input masks utilized by all models, while the same position in our results highlights the A-O-DisCo required by our approach. The blue arrow indicates the source region for the magnified view presented in the first row.}
    \label{fig:outpainting_light}
    \vspace{-0.1in}
\end{figure*}

\vspace{0.5em}
\noindent\textbf{Outpainting.}~For evaluation, we outpaint videos from $280\times 520$ to $480 \times 720$. As demonstrated in~\cref{tab:comparison} (b), O-DisCo-Edit establishes new SOTA records across all metrics. VACE generates grainy textures in edited areas, Senorita and
VideoPainter  exhibit noticeable box-like artifacts  shown in~\cref{fig:outpainting_light}. In contrast, O-DisCo-Edit creates exceptionally well-blended, natural, and continuous results.

\vspace{0.5em}
\noindent\textbf{Object Internal Motion Transfer.}~Existing control signals struggle to accurately capture intricate object internal motions,  such as the milk flow depicted in the~\cref{fig:remove_style_inside_swap}.  Our method addresses this by leveraging the masks of the bottle and cup to generate a corresponding  A-O-DisCo,  thereby achieving precise transfer of internal object motion.  In contrast,  other baseline approaches are unable to accomplish accurate internal motion transfer.  As shown in the quantitative results in~\cref{tab:comparison} (c),  we achieve the best overall metrics.  Consequently,  our method yields the most superior internal motion transfer results.

\vspace{0.5em}
\noindent\textbf{Lighting  Transfer.}~O-DisCo-Edit is capable of simultaneously editing objects and transferring their lighting transformation.  Quantitative results,  presented in~\cref{tab:comparison} (d),  demonstrate that we achieve the best overall metrics.  Qualitatively,  as illustrated in~\cref{fig:Add_remove_change_color},  the original image exhibits obviously lighting and shadow changes.  All other baseline methods fail to transfer these variations,  whereas our approach successfully achieves excellent transfer performance.  Consequently,  our method yields the most superior lighting and shadow transfer results.

\begin{table*}[htbp]
  \centering
       \setlength{\tabcolsep}{2mm}
\caption{Ablation Studies on swap and object removal task. \normalsize{\textcircled{\scriptsize{1}}}  CFP module,~\normalsize{\textcircled{\scriptsize{2}}}~A-O-DisCo,~\normalsize{\textcircled{\scriptsize{3}}} IDP module. 
    ``w/o \normalsize{\textcircled{\scriptsize{1}}}\normalsize \normalsize{\textcircled{\scriptsize{2}}}\normalsize \normalsize{\textcircled{\scriptsize{3}}}'' denotes training with R-O-DisCo and inference with a fixed signal, entirely omitting IDP and CFP modules.  ``w/o \normalsize{\textcircled{\scriptsize{2}}}\normalsize \normalsize{\textcircled{\scriptsize{3}}}'' indicates training without module IDP and inference with a fixed signal. ``w/o \normalsize{\textcircled{\scriptsize{2}}}'' refers to using a fixed inference signal.  ``w/o \normalsize{\textcircled{\scriptsize{3}}}'' indicates training without the IDP module. “TC” denotes temporal consistency.}
  \setlength{\tabcolsep}{0.8mm}
  \resizebox{1\textwidth}{!}{
    \begin{tabular}{l|cccccc|cccccl}
    \toprule
    Metric & \multicolumn{2}{c}{Video Quality} & \multicolumn{2}{c}{Alignment} & \multicolumn{2}{c}{Preservation } & \multicolumn{2}{c}{Video Quality} & \multicolumn{2}{c}{Removal Capability} & \multicolumn{2}{c}{Preservation} \\
    \midrule
    Model & TC$\uparrow$    & FVD$\downarrow$   & CLIP-T$\uparrow$ & CLIP-I$_\text{E}$$\uparrow$ & PSNR$_\text{P}$$\uparrow$ & SSIM$_\text{P}$$\uparrow$ & TC$\uparrow$ & FVD$\downarrow$ & PSNR$_\text{E}$$\uparrow$ & {SSIM$_\text{E}$}$\uparrow$ & {PSNR$_\text{P}$}$\uparrow$ & {SSIM$_\text{P}$}$\uparrow$\\
    \midrule
    & \multicolumn{6}{l|}{\textbf{Task: Swap}} & \multicolumn{6}{l}{\textbf{Task:Object Removal (33)}} \\
    {w/o \normalsize{\textcircled{\scriptsize{1}}}\normalsize \normalsize{\textcircled{\scriptsize{2}}}\normalsize \normalsize{\textcircled{\scriptsize{3}}}\normalsize} & 0.9837 & 887.6 & \textbf{15.92} & \textbf{92.13} & 21.57 & 0.6534 & \textbf{0.9883} & 2428  & 22.30  & 0.9164 & 25.41 & 0.8195 \\
    {w/o \normalsize{\textcircled{\scriptsize{2}}}\normalsize \normalsize{\textcircled{\scriptsize{3}}}\normalsize} & 0.9834 & 866.3 & \underline{15.75} & 91.03 & 25.32 & 0.8037 & \underline{0.9882} & 2167  & 20.80  & 0.9106 & 27.17 & 0.8913 \\
    w/o \normalsize{\textcircled{\scriptsize{2}}}\normalsize & \underline{0.9839} & \underline{776.8} & 15.25 & 91.69 & \underline{26.09} & \underline{0.8087} & \underline{0.9882} & 2103  & 21.44 & 0.9254 & 27.69 & \underline{0.8942} \\
    w/o \normalsize{\textcircled{\scriptsize{3}}}\normalsize & \underline{0.9839} & 796.7 & 15.41 & 91.60  & 25.47 & 0.8053  & 0.9968 & \underline{362.8} & \underline{35.90}  & \underline{0.9737} & \underline{29.10}  & \textbf{0.9029} \\
    \cellcolor[rgb]{ .91,  .91,  .91}O-DisCo-Edit & \cellcolor[rgb]{ .91,  .91,  .91}\textbf{0.9840} & \cellcolor[rgb]{ .91,  .91,  .91}\textbf{711.8} & \cellcolor[rgb]{ .91,  .91,  .91}15.27 & \cellcolor[rgb]{ .91,  .91,  .91}\underline{91.84} & \cellcolor[rgb]{ .91,  .91,  .91}\textbf{26.25} & \cellcolor[rgb]{ .91,  .91,  .91}\textbf{0.8096}  & \cellcolor[rgb]{ .91,  .91,  .91}0.9969 & \cellcolor[rgb]{ .91,  .91,  .91}\textbf{360.1} & \cellcolor[rgb]{ .91,  .91,  .91}\textbf{36.05} & \cellcolor[rgb]{ .91,  .91,  .91}\textbf{0.9740} & \cellcolor[rgb]{ .91,  .91,  .91}\textbf{29.11} & \cellcolor[rgb]{ .91,  .91,  .91}\textbf{0.9029}\\
    \bottomrule
    \end{tabular}}%
   \vspace{-1em} 
  \label{tab:Ablation}%
\end{table*}%

\vspace{0.5em}
\noindent\textbf{Color Change.}~O-DisCo-Edit is capable of  changing color while preserving intrinsic characteristics, an advantage supported by both quantitative and qualitative results.  
In quantitative analysis, our approach achieves the highest average score as shown in~\cref{tab:comparison} (e). 
Qualitatively, a visual comparison in~\cref{fig:Add_remove_change_color} reveals that VACE produces irregular color gradients, while Senorita and VideoPainter generate subtle artifacts. 
Therefore, our approach avoids these issues and yields superior color transformation results. 
Notably, Senorita's top score in user studies comes from its first-frame propagation strategy.  While this strategy creates high visual consistency, it does so at the expense of poor preservation in non-edited regions.

\vspace{0.5em}
\noindent\textbf{Swap.}~Quantitative evaluation in~\cref{tab:comparison} (f) shows O-DisCo-Edit's performance is second to VACE, yet we achieve a higher $\text{CLIP-I}_\text{E}$. 
As shown in~\cref{fig:remove_style_inside_swap}, VACE 14B struggles with ID consistency. Meanwhile, VACE 1.3B and VideoPainter overfit masks boundaries, generating anatomically incorrect outputs (e.g., polar bears with three ears). 
Furthermore, Senorita, VACE 14B, and VACE 1.3B  exhibit motion inconsistencies (red box). 
Conversely, our method exhibits superior visual results, as evidenced by user study.

\vspace{0.5em}
\noindent\textbf{Addition.}~O-DisCo-Edit enables adding new objects to existing moving objects in a video. 
As shown in~\cref{tab:comparison} (g), O-DisCo-Edit reaches competitive performance, ranking second only to Senorita. 
However, as~\cref{fig:Add_remove_change_color} illustrated, Senorita fail to complete the addition task, with its high metrics solely due to ``copying'' the original video. Therefore, our method attains the most preferred additions results in user study.

\begin{table}[htbp]
\centering
\setlength{\tabcolsep}{1mm}
\small
\caption{Comparison of different models' performance metrics on style transfer. ``Preservation'' means  non-edited region preservation, ``TC'' denotes temporal consistency, ``EC'' represents editing completeness, and ``VQ'' stands for visual quality.}
\begin{tabular}{lcccccc}
\toprule
 Metrics &  \multicolumn{2}{c}{Video Quality} & Preservation & \multicolumn{2}{c}{User Study}  \\
    \midrule
Method & TC$\uparrow$  & ArtFID$\downarrow$  & CFSD$\downarrow$  & EC$\uparrow$     & VQ$\uparrow$\\
\midrule
Senorita2m & \textbf{0.9960} & 7.979  & \textbf{0.0933} & 2.989 & 2.578  \\
O-DisCo-Edit & 0.9954 & \textbf{7.292}  & 0.2055 & \textbf{4.322} & \textbf{4.156} \\
\bottomrule
\end{tabular}

  \label{tab:style_transfer}
  \vspace{-1em}
\end{table}
\vspace{0.5em}
\noindent\textbf{Style Transfer.}~O-DisCo-Edit attains the highest ArtFID, as shown in~\cref{tab:style_transfer}. 
In contrast, Senorita exhibits a very low CFSD, which indicates a tendency for its generated videos to align with the original reference content. As depicted in~\cref{fig:remove_style_inside_swap}, such alignment is detrimental to style transfer quality. Therefore, our method consistently received higher user study evaluations.


\subsection{Ablation Analysis}
As shown in~\cref{tab:Ablation}, we ablate on O-DisCo-Edit.  (1) Comparing row 1 and row 2,  a significant improvement in both PSNR$_\text{P}$ and SSIM$_\text{P}$ is observed with the inclusion of the CFP module.  Thus this results validate CFP module's effectiveness in  non-edited regions preservation. 
(2) When contrasting row 3/4 with row 2, the A-O-DisCo and IDP modules individually enhance the generated video quality and the preservation of non-edited regions for both tasks. Additionally, they improve the appearance consistency (CLIP-I$_\text{E}$) of the edited regions for the swap task and the removal capability for the object removal task.
(3) A further comparison between rows 3/4 and row 5 reveals that the combination of the A-O-DisCo and IDP modules leads to an even greater improvement in model performance.

\section{Conclusion}
\label{sec:conclusion}
In this work, we introduced O-DisCo-Edit, a unified framework designed to address the key challenges in controllable video editing. Our core innovation, O-DisCo, unifies various editing signals into a single, noise-based representation. This not only dramatically simplifies the training process and reduces resource demands but also enables  multi-granularity editing during inference. Paired with the designed CFP module, O-DisCo-Edit can accomplish high-fidelity editing while robustly preserving unedited regions.
Comprehensive experiments on eight different tasks show that O-DisCo-Edit achieves new SOTA results, outperforming both specialized and multi-task models.
This success offers a new perspective on  video editing research, that a single unified control signal can be both versatile and precise without sacrificing efficiency.

\newpage
{
    \small
    \bibliographystyle{unsrt}
    \bibliography{main}
}

\clearpage
\appendix
\section*{Appendix}

\section{More Details about Methodology}
\subsection{Random Object Distortion Control}
During the training phase, we employ a random distorter to generate the random object distortion control (R-O-DisCo), as detailed in Algorithm \ref{alg:random_distorter}. The algorithm takes the reference video $V_\text{ref}$ and mask $M$ as input, and outputs the R-O-DisCo, $V_\text{RODC}$. The process begins by randomly sampling several distortion parameters from uniform distributions: a scaling factor ($\theta$) from $[1.5, 3.0]$, a scaling mode ($\mu$) to determine whether to scale up or down, a target channel ($c^*$)  is the scaled channel, a color offset ($\delta$) from $\{-100, -50, 50, 100\}$, and a block size ($b$) for mosaicking from $\{8, 10, 12, 15, 16, 20, 24\}$. The functions used in this process are defined as follows: ZerosLike() returns a tensor of zeros with the same shape as the input video; Clip() constrains the input values to the range $[0,255]$; AvgPool() performs downsampling using average pooling; and Interpolate() performs upsampling using nearest-neighbor interpolation. Based on these, the distorter applies both color distortion and mosaicking to the reference video. The final  $V_\text{RODC}$, is then obtained by using the mask $M$ to apply these distortions exclusively to the  object that needs to be edited.

\subsection{Adaptive Object Distortion Control}
During inference phase,  our model adapts to specific tasks or instructions via adaptive object distortion control (A-O-DisCo), which is implemented by an adaptive distorter.   
It is achieved through contrast modification (scaling and clipping) and dynamic noise injection within the  editable regions. 
The adaptive controller determines suitable values for contrast parameter $\alpha$, noise intensity $\sigma$,  and normalized Gaussian kernel size $k$ by calculating two similarities: $\mathbf{Sim}_{i}$,  the edited region's edge map similarity between the reference image and the reference video's first frame; $\mathbf{Sim}_\text{v}$, the intra-frame similarity within the reference video's edited region edge map. Empirically, fitting these three parameters using a quadratic polynomial of two similarity  yields superior results. 

In the Algorithm \ref{alg:adaptive_tracking}, $V_\text{ref}$ is the reference video, $M$ represents the mask, $I_\text{ref}$ is the reference image, and $T=49$ is the number of video frames. The functions are defined as follows: Canny() computes the edge map; SSIM$( , | f_\text{masks})$ calculates the SSIM value on the edited region, which is  extracted using the $f_\text{masks}$; Odd() returns the nearest odd integer value; and GaussianBlur() performs a Gaussian blur. The parameters are defined by the following quadratic functions: $f_1=3000 \cdot {\mathbf{Sim}_{i}}^2 +6000 \cdot {\mathbf{Sim}_{i}} + 300$, $f_2=4622.64 \cdot {\mathbf{Sim}_\text{v}}^2 + 92453.28\cdot {\mathbf{Sim}_\text{v}}+4623.64 $, and $f_3=-36\cdot {\mathbf{Sim}_\text{v}}^2+72 \cdot {\mathbf{Sim}_\text{v}}-35 $. The process involves two passes: in the first, we iterate through each video frame, calculate the similarity metrics (using SSIM), and determine the adaptive parameters; in the second, we apply scaling \& clipping and Gaussian blurring to the reference video frames. Finally, using a mask, we obtain the A-O-DisCo video, $V_\text{AODC}$.

\begin{algorithm}[tb]
\caption{Random Distorter}
\label{alg:random_distorter}
\textbf{Input}: $V_\text{ref}$, $M$ \\
\textbf{Output}: $V_\text{RODC}$
\begin{algorithmic}[1]
\STATE Initialize processing parameters:
\STATE \quad Scaling factor: $\theta \sim {U}(1.5, 3.0)$
\STATE \quad Target channel: $c^* \sim {U}(\{0,1,2\})$
\STATE \quad Color offset: $\delta \sim {U}(\{-100,-50,50,100\})$
\STATE \quad Block size: $b \sim {U}(\{8,10,12,15,16,20,24\})$
\STATE \quad Scaling mode: $\mu \sim {U}(\{0,1\})$
\STATE Initialize $V_\text{cd} \gets \text{ZerosLike}(V_\text{ref})$
\STATE Color Distortion
\FOR{each channel $c \in \{0,1,2\}$}
    \IF{$c = c^*$}
        \IF{$\mu = 0$}
            \STATE $V_\text{cd}[:,:,:,c] \gets \text{Clip}(V_\text{ref}[:,:,:,c] \cdot \theta, 0, 255)$
        \ELSE
            \STATE $V_\text{cd}[:,:,:,c] \gets \text{Clip}(V_\text{ref}[:,:,:,c] / \theta, 0, 255)$
        \ENDIF
    \ELSIF{$c = 0$}
        \STATE $V_\text{cd}[:,:,:,c]  \gets \text{Clip}(V_\text{ref}[:,:,:,c]  + \delta, 0, 255)$
    \ELSE
        \STATE $V_\text{cd}[:,:,:,c]  \gets \text{Clip}(V_\text{ref}[:,:,:,c]  - \delta, 0, 255)$
    \ENDIF
\ENDFOR
\STATE Mosaicking
\STATE $V_\text{low} \gets \text{AvgPool}(V_\text{cd}, b)$
\STATE $V_\text{cdm} \gets \text{Interpolate}(V_\text{low}, b)$

\STATE $V_\text{RODC} \gets V_\text{ref} \odot (\mathbf{1} - M) + V_\text{cdm} \odot M$
\STATE \textbf{return} $V_\text{RODC}$ 
\end{algorithmic}
\end{algorithm}

\begin{algorithm}[tb]
\caption{Adaptive Distorter}
\label{alg:adaptive_tracking}
\textbf{Input}: $V_\text{ref}$, $M$, $I_\text{ref}$\\
\textbf{Output}: $V_\text{AODC}$
\begin{algorithmic}[1]
\STATE \textbf{First Pass: Calculate SSIM Metrics} 
\FOR{$t \gets 0$ \TO $T-1$}
    \STATE $f_\text{ref} \gets V_\text{ref}[t]$ 
    \STATE $f_\text{masks} \gets M[t]$ 
    \IF{$t=0$}
        \STATE $f_\text{ref}^\text{canny}  \gets \text{Canny}(f_\text{ref}) $
        \STATE $I_\text{ref}^\text{canny}  \gets \text{Canny}(I_\text{ref}) $
        \STATE $f_\text{old} \gets f_\text{ref}^\text{canny} $
        \STATE $\mathbf{Sim}_{i} \gets \text{SSIM}(f_\text{ref}^\text{canny}, I_\text{ref}^\text{canny} | f_\text{masks})$
    \ELSE
        \STATE $f_\text{ref}^\text{canny}  \gets \text{Canny}(f_\text{ref}) $
        \STATE $f_\text{new}  \gets f_\text{ref}^\text{canny} $
        \STATE $ssim_\text{v}[t] \gets \text{SSIM}(f_\text{new}, f_\text{old} | f_\text{masks})$
        \STATE $f_\text{old} \gets f_\text{ref}^\text{canny} $
    \ENDIF
\ENDFOR
\STATE Compute average SSIM of $ssim_\text{v}[t]$, to obtain $\mathbf{Sim}_\text{v}$
\STATE \textbf{Calculate Adaptive Parameters}
\STATE $k \gets 0.2 \cdot \text{Odd}(f_1(\mathbf{Sim}_{i}) + 1.2 \cdot f_2(\mathbf{Sim}_\text{v}))$
\STATE $\sigma \gets 0.2 \cdot (f_1(\mathbf{Sim}_{i}) + 1.2 \cdot f_2(\mathbf{Sim}_\text{v}))$
\STATE $\alpha \gets f_3(\mathbf{Sim}_\text{v})$
\STATE \textbf{Second Pass: Get A-O-DisCo}
\STATE Initialize $V_\text{AODC} \gets \text{ZerosLike}(V_\text{ref})$
\FOR{$t \gets 0$ \TO $T-1$}
    \STATE $f_\text{ref} \gets V_\text{ref}[t]$ 
    \STATE $f_\text{masks} \gets M[t]$ 
    \STATE $f_\text{AODC} \gets V_\text{AODC}[t]$ 
    \STATE Apply adaptive transformations:
    \STATE $f_\text{c} \gets \text{Clip}(\alpha \cdot V_\text{ref}, 0, 255)$
    \STATE $f_\text{cn} \gets \text{GaussianBlur}(f_\text{ref}, 0, 255, k, \sigma)$
    \STATE $f_\text{AODC} \gets f_\text{ref} \odot (\mathbf{1} - M) + V_\text{cn} \odot M$
\ENDFOR
\STATE \textbf{return} $V_\text{AODC}$
\end{algorithmic}
\end{algorithm}

\subsection{``Copy-Form''  Preservation Module}
We propose the copy-form preservation (CFP) module, which achieves preservation of non-edited regions with an effect similar to ``first-frame copying". The CFP module is expressed as:
\begin{equation}
\begin{aligned}
\label{copy_form}
     & z^{\text{v}'}_\text{p}=z^\text{v}_\text{ref} \odot (\mathbf{1}-z_\text{mask}[1:]), \\
     & z_\text{images} =[z^\text{i}_\text{ref}, z^{\text{v}'}_\text{p}],
\end{aligned}
\end{equation}

During training, we randomly dilate the masks $z_\text{mask}[1:]$ using a max pooling operation, where the kernel size is uniformly sampled from the set $\{1, 3, 5, 7, 9, 11, 13, 15, 17, 19, 21\}$ with equal probability. During inference, users can adaptively adjust the dilation kernel size based on the target object's size in the reference image. This allows for fine-tuning the edited region in the reference video to accommodate objects of different sizes for tasks such as swap or addition.

\section{More Details about Experiments}
\subsection{Construction of the benchmark}
Video editing is a rapidly advancing field. However, a public benchmark for first-frame-based video editing is still lacking. While the widely used DIVAS~\citep{pont20172017}  is available, its lack of edited images makes it unsuitable for evaluating video editing models. The VACE benchmark~\citep{jiang2025vace}, on the other hand, only openly provides 1-2 examples per task, which is insufficient for a comprehensive and reliable evaluation. The Genprop-Data benchmark~\citep{liu2025generative} contains 40 video-image pairs, but its quality is compromised by videos with variable frame counts and dimensions. Most notably, the edited images in this benchmark exhibit significant background discrepancies compared to the original videos. Although models like OmniV2V~\citep{liang2025omniv2v} and UNIC~\citep{ye2025unic} have constructed their own benchmarks, none of them are publicly available.

Given the scarcity of public video editing benchmarks, we construct a new multi-task video editing benchmark to comprehensively evaluate the performance of our method across various tasks, which includes tasks such as outpainting, object internal motion transfer, lighting transfer, color change, swap, addition, and style transfer. The data is sourced from DAVIS~\citep{pont20172017} and VPData~\citep{bian2025videopainter}. The specific construction process is as follows:
\begin{itemize}
\item For outpainting: We select  videos containing a variety of natural landscapes (mountains, lakes, oceans, skies) and man-made scenes (villages, buildings, roads) to ensure diversity.
\item For object internal motion transfer: We choose  videos with clear internal motion, such as water flowing inside a bottle, a rotating sphere, and the movement of a clock's hands. To specifically highlight the effect of internal motion transfer, we edit the first frame using relatively simple color and swap instructions.
\item For lighting transfer: We select  videos with significant lighting changes, including the play of light on water surfaces or metal, and lighting variations for both underwater and terrestrial object. Similarly, to focus on the lighting transfer effect, we edit the first frame with simple color and swap instructions.
\item For other tasks:  We randomly select several videos. We then use Qwen-VL2.5 7B~\citep{bai2023qwen} to randomly generate editing instructions. We use HiDream~\citep{cai2025hidream} and commercial models~\footref{fn:label1} to generate the edited first frame. From these, we select the best-edited first frames as reference images and their corresponding videos as reference videos. To increase the complexity of the swap and addition tasks, we further filter for videos that included obvious and complex motion.
\end{itemize}

In total, we collect  a benchmark of 134 sets, each containing a reference video, a reference image, and a mask video.  All of them have a resolution of $480\times720$ and video consist of 49 frames.
\begin{table}[htbp]
  \centering
  \caption{A comparison between our method and specialized models on object removal task (49). The evaluation is based on several key metrics, including DBR (degree of object removal), BC (background consistency), VQ (video quality), and EC (editing completeness).}
  \setlength{\tabcolsep}{1mm}
    \begin{tabular}{l|cccc}
    \toprule
    Method & {Diffuearser} & {Minimax} & {Propainter} & {\cellcolor[rgb]{ .949,  .949,  .949}O-DisCo-Edit} \\
    \midrule
    DRO & 3.356  & 3.800  & 3.311  & \cellcolor[rgb]{ .949,  .949,  .949}\textbf{4.244 } \\
    BC & 2.889  & 3.333  & 2.778  & \cellcolor[rgb]{ .949,  .949,  .949}\textbf{3.689 } \\
    EC    & 3.122  & 3.567  & 3.044  & \cellcolor[rgb]{ .949,  .949,  .949}\textbf{3.967 } \\
    VQ & 2.867  & 3.333  & 2.822  & \cellcolor[rgb]{ .949,  .949,  .949}\textbf{3.689 } \\
    \bottomrule
    \end{tabular}%
    
  \label{tab:Object_Removal_49}%
  \vspace{-1em}
\end{table}%
Additionally, we evaluate the object removal task using the OmnimatteRF benchmark~\citep{lin2023omnimatterf}. Since ground truth background videos for removal are often unavailable in real-world scenarios, we use a commercial model~\footref{fn:label1} to perform object removal on the first frame of the reference video. For a fairer comparison, this processed first frame is used as the input for multi-task baselines.

\subsection{Choice of Evaluation Metrics}
For the color change, object internal motion transfer, and lighting transfer tasks, we choose ArtFID~\citep{wright2022artfid} as our video quality metric. As the author of~\citep{chung2024style} notes, ArtFID is calculated as $(1+\text{LPIPS})\cdot(1+\text{FID})$. Specifically, LPIPS~\citep{zhang2018unreasonable} measures content fidelity between the generated video frames and the reference video frames, while FID~\citep{heusel2017gans} assesses style fidelity between the generated video frames and the corresponding edited first frame.

Just as described in the benchmark construction, the color change, object internal motion transfer, and lighting transfer tasks are analogous to local style transfer. ArtFID is a better metric for these tasks because it can tell the difference between content fidelity and style fidelity. This gives it a more detailed understanding of video quality than FVD, which only measures the overall video distribution. For this reason, ArtFID is especially well-suited for evaluating these specific tasks.

\begin{table*}[htbp]
  \centering
      \caption{This table presents a comparison of user study results for different models. The metrics used for evaluation include: DBR (degree of object removal), BC (background consistency), VQ (video quality), EC (editing completeness), OE (outpainting effectiveness), IMC (internal motion consistency), IDC (ID consistency), LC (lighting consistency), ColorC (color consistency), COCC (consistency of non-color content), MC (motion consistency), PAOM (plausibility of the added object’s motion), CNER (consistency of non-edited regions), CC (content consistency), and SC (style consistency). }
    \begin{tabular}{cc|ccccc}
    \toprule
    Task  & Metric & VACE 1.3B & VACE 14B & Senorita & VideoPainter & \cellcolor[rgb]{ .91,  .91,  .91}O-DisCo-Edit \\
    \midrule
    \multirow{4}[2]{*}{\makecell{Object \\{}Removal (33)}} & DOR   & 2.200  & 1.622  & 3.511  & 1.844  & \cellcolor[rgb]{ .91,  .91,  .91}\textbf{4.311 } \\
          & BC    & 1.822  & 1.533  & 3.111  & 1.644  & \cellcolor[rgb]{ .91,  .91,  .91}\textbf{3.600 } \\
          & EC    & 2.011  & 1.578  & 3.311  & 1.744  & \cellcolor[rgb]{ .91,  .91,  .91}\textbf{3.956 } \\
          & VQ    & 1.911  & 1.444  & 3.156  & 1.578  & \cellcolor[rgb]{ .91,  .91,  .91}\textbf{3.689 } \\
    \midrule
    \multirow{3}[2]{*}{Outpainting} & OE    & 4.244  & 4.178  & 2.889  & 2.156  & \cellcolor[rgb]{ .91,  .91,  .91}\textbf{4.289 } \\
          & EC    & 4.244  & 4.178  & 2.889  & 2.156  & \cellcolor[rgb]{ .91,  .91,  .91}\textbf{4.289 } \\
          & VQ    & 3.933  & 3.911  & 2.600  & 1.978  & \cellcolor[rgb]{ .91,  .91,  .91}\textbf{4.067 } \\
    \midrule
    \multirow{4}[2]{*}{\makecell{Object Internal\\ Motion Transfer}} & IMC   & 2.956  & 3.333  & 2.511  & 2.156  & \cellcolor[rgb]{ .91,  .91,  .91}\textbf{4.267 } \\
          & IDC   & 3.067  & 2.911  & 3.533  & 2.444  & \cellcolor[rgb]{ .91,  .91,  .91}\textbf{4.089 } \\
          & EC    & 3.011  & 3.122  & 3.022  & 2.300  & \cellcolor[rgb]{ .91,  .91,  .91}\textbf{4.178 } \\
          & VQ    & 2.533  & 2.800  & 2.333  & 1.822  & \cellcolor[rgb]{ .91,  .91,  .91}\textbf{3.756 } \\
    \midrule
    \multirow{4}[2]{*}{\makecell{Light \\Transfer}} & LC    & 2.867  & 3.333  & 2.867  & 2.800  & \cellcolor[rgb]{ .91,  .91,  .91}\textbf{4.200 } \\
          & IDC   & 3.267  & 3.489  & 3.200  & 3.022  & \cellcolor[rgb]{ .91,  .91,  .91}\textbf{3.756 } \\
          & EC    & 3.067  & 3.411  & 3.033  & 2.911  & \cellcolor[rgb]{ .91,  .91,  .91}\textbf{3.978 } \\
          & VQ    & 2.644  & 3.067  & 2.400  & 2.489  & \cellcolor[rgb]{ .91,  .91,  .91}\textbf{3.689 } \\
    \midrule
    \multirow{4}[2]{*}{\makecell{Color  \\ Change}} & ColorC & 3.644  & 3.378  & \textbf{4.244 } & 3.778  & \cellcolor[rgb]{ .91,  .91,  .91}4.133  \\
          & COCC  & 3.622  & 3.533  & \textbf{3.822 } & 3.489  & \cellcolor[rgb]{ .91,  .91,  .91}3.756  \\
          & EC    & 3.633  & 3.456  & \textbf{4.033 } & 3.633  & \cellcolor[rgb]{ .91,  .91,  .91}3.944  \\
          & VQ    & 3.244  & 3.089  & \textbf{3.711 } & 3.267  & \cellcolor[rgb]{ .91,  .91,  .91}3.689  \\
    \midrule
    \multirow{4}[2]{*}{Swap} & MC    & 3.711  & 3.911  & 3.533  & 3.489  & \cellcolor[rgb]{ .91,  .91,  .91}\textbf{3.956 } \\
          & IDC   & 3.222  & 3.200  & 3.378  & 2.444  & \cellcolor[rgb]{ .91,  .91,  .91}\textbf{4.111 } \\
          & EC    & 3.467  & 3.556  & 3.456  & 2.967  & \cellcolor[rgb]{ .91,  .91,  .91}\textbf{4.033 } \\
          & VQ    & 2.956  & 3.089  & 2.956  & 2.178  & \cellcolor[rgb]{ .91,  .91,  .91}\textbf{3.689 } \\
    \midrule
    \multirow{5}[2]{*}{Addtion} & PAOM  & 3.289  & 3.200  & 3.222  & 3.178  & \cellcolor[rgb]{ .91,  .91,  .91}\textbf{4.111 } \\
          & IDC   & 3.267  & 2.711  & 3.267  & 3.067  & \cellcolor[rgb]{ .91,  .91,  .91}\textbf{4.267 } \\
          & CNER  & 3.556  & 3.689  & 4.000  & 3.067  & \cellcolor[rgb]{ .91,  .91,  .91}\textbf{3.733 } \\
          & EC    & 3.370  & 3.200  & 3.496  & 3.104  & \cellcolor[rgb]{ .91,  .91,  .91}\textbf{4.037 } \\
          & VQ    & 3.089  & 2.578  & 2.911  & 2.489  & \cellcolor[rgb]{ .91,  .91,  .91}\textbf{3.822 } \\
    \midrule
    \multirow{4}[2]{*}{\makecell{Style\\ Transfer}} & SC    & \textbackslash{} & \textbackslash{} & 2.911  & \textbackslash{} & \cellcolor[rgb]{ .91,  .91,  .91}\textbf{4.356 } \\
          & CC    & \textbackslash{} & \textbackslash{} & 3.067  & \textbackslash{} & \cellcolor[rgb]{ .91,  .91,  .91}\textbf{4.289 } \\
          & EC    & \textbackslash{} & \textbackslash{} & 2.989  & \textbackslash{} & \cellcolor[rgb]{ .91,  .91,  .91}\textbf{4.322 } \\
          & VQ    & \textbackslash{} & \textbackslash{} & 2.578  & \textbackslash{} & \cellcolor[rgb]{ .91,  .91,  .91}\textbf{4.156 } \\
    \bottomrule
    \end{tabular}%
    \vspace{-1em}    
  \label{tab:human_eval_detais}%
\end{table*}%

\subsection{User Study}
We conduct a user study where anonymized generated videos are randomly distributed to participants for scoring on a 1-5 scale. For each task, we define specific evaluation criteria:
\begin{itemize}
\item For object removal: Users score the generated videos on degree of object removal (DOR), background consistency (BC) with the background video, and overall video quality (VQ).
\item Outpainting: Users evaluate outpainting effectiveness (OE) and overall video quality (VQ).
\item Object internal motion transfer: The criteria are the internal motion consistency (IMC) between the generated and reference videos, ID consistency (IDC) between the generated video and reference image, and overall video quality (VQ).
\begin{figure}[!t]
\centering
\includegraphics[width=0.48\textwidth]{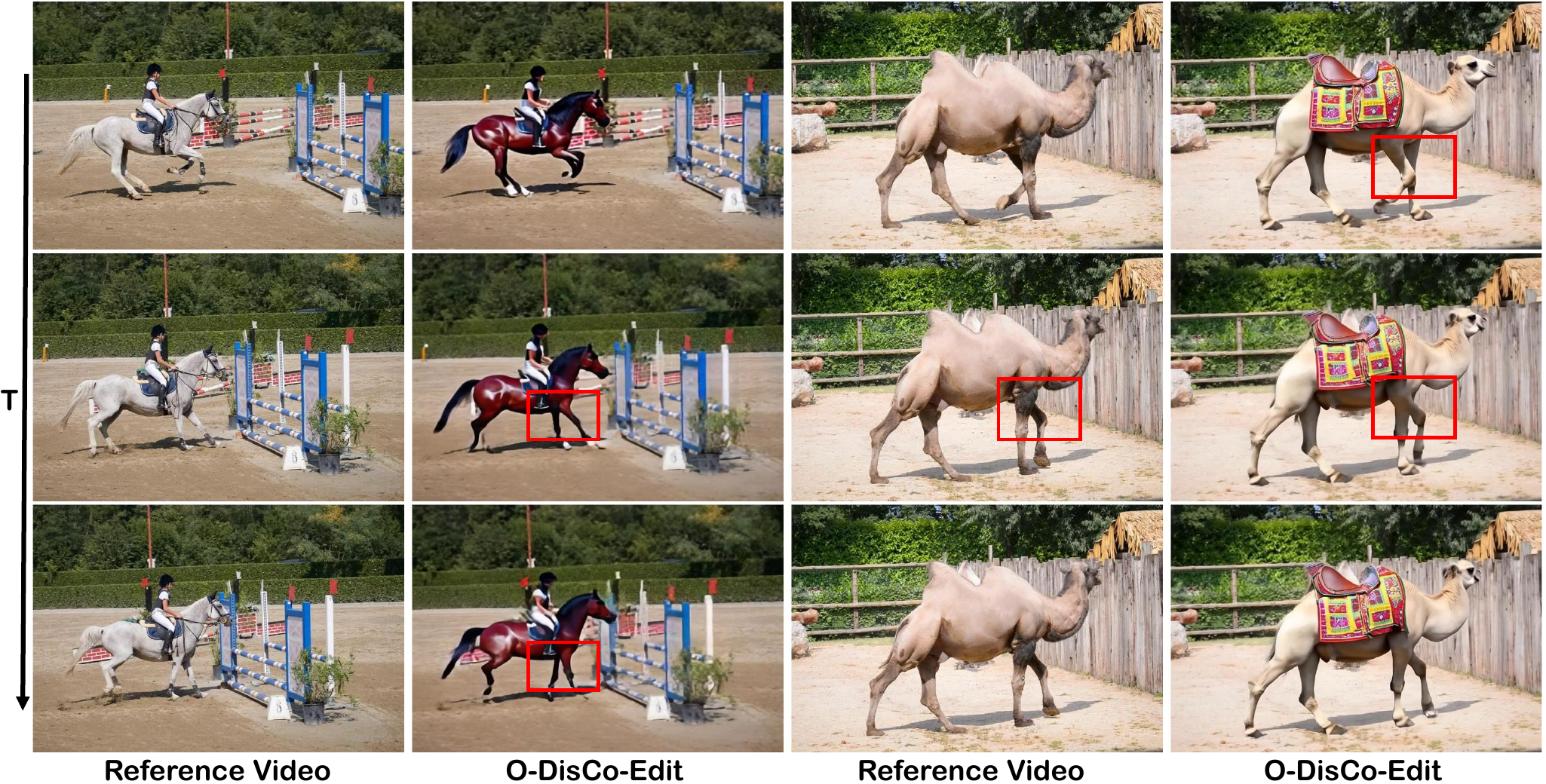}
\caption{The bad cases of O-DisCo-Edit, where ``T'' denotes the temporal axis.}
\vspace{-1em}
\label{fig:Bad_case}
\end{figure}

\item Lighting transfer: Users score lighting consistency (LC) between the generated and reference videos, ID consistency (IDC) with the reference image, and overall video quality (VQ).
\item Color change: User evaluate color consistency (ColorC) with the reference image, consistency of non-color content (COCC) with the reference video, and overall video quality (VQ).
\item Swap: The criteria include ID consistency (IDC) with the reference image, motion consistency (MC) with the reference video, and overall video quality (VQ).
\item Addition: Users score ID consistency (IDC) with the reference image, the plausibility of the added object's motion (PAOM), the consistency of non-edited regions (CNER) with the reference video, and overall video quality (VQ).
\item Style transfer: We define criteria for content consistency (CC) with the reference video, style consistency (SC) with the reference image, and overall video quality (VQ).
\end{itemize}

In our user study, we follow the methodology of~\citep{ye2025imgedit} to ensure that editing effectiveness was prioritized. We do this by capping the video quality score at the editing effectiveness score for each submission, meaning the video's overall quality score cannot exceed its score for how well the edit is performed. For example, in the object removal task, the video quality score is capped by the degree of object removal.  We then compute the overall editing completeness (EC) as the average of all metrics except for video quality. In total, we collecte 9 valid responses for the user study. The detailed results of this user study are presented in~\cref{tab:Object_Removal_49} and~\cref{tab:human_eval_detais}. Our method achieve the best user study results across all tasks except for color change. Notably, Senorita's top score in user studies comes from its first-frame propagation strategy. As discussed in the main submission, this strategy creates high visual consistency but performs poorly at preserving non-edited regions.

\subsection{Bad Case Analysis}
As revealed by~\cref{tab:comparison}, our model receive its lowest scores on the swap task compared to its performance on other tasks. We attribute this poor performance to our O-DisCo-Edit model's difficulty in handling complex, four-limbed object motions, as illustrated by the misaligned  legs of the swapped object in~\cref{fig:Bad_case}. We hypothesize this issue stems from three potential factors. First, the base model may have inherent limitations; as shown in~\cref{fig:Bad_case_ana}, Senorita~\citep{zi2025se} and VideoPainter~\citep{bian2025videopainter}, which share our base model, exhibit similar misaligned leg problems. In contrast, while VACE 14B~\citep{jiang2025vace} shows some ID distortion, its legs maintain correct depth and position. Second, the model's parameter count may be a factor, as VACE 1.3B (a smaller version of VACE 14B) also displays this failure mode. This suggests that increasing our model's parameter count could potentially resolve the issue. Finally, the training dataset we used contains a limited number of examples with complex four-limbed animal motions, which likely prevents the model from generalizing well to such challenging cases.

\begin{figure*}[!t]
\centering
\includegraphics[width=1\textwidth]{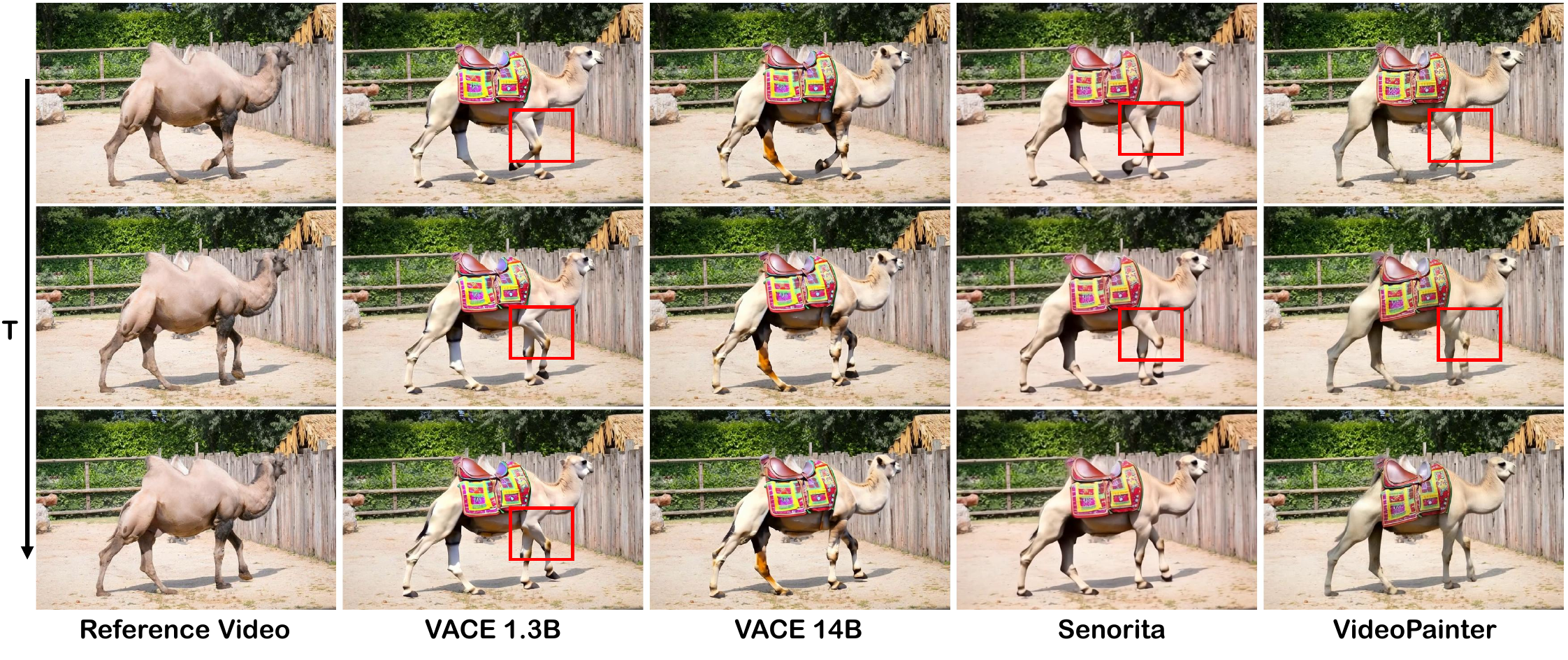}
\caption{Qualitative results of the baseline models on swap task involving complex, four-limbed object motions. The temporal axis is denoted by 'T'.}
\label{fig:Bad_case_ana}
\end{figure*}

\subsection{More Ablation Experiment}
\begin{table}[htbp]
  \centering
  \caption{The results of the ablation experiment  on  outpainting task. \normalsize{\textcircled{\scriptsize{1}}}  CFP module,~\normalsize{\textcircled{\scriptsize{2}}}~A-O-DisCo,~\normalsize{\textcircled{\scriptsize{3}}} IDP module. 
    ``w/o \normalsize{\textcircled{\scriptsize{1}}}\normalsize \normalsize{\textcircled{\scriptsize{2}}}\normalsize \normalsize{\textcircled{\scriptsize{3}}}'' denotes training with R-O-DisCo and inference with a fixed signal, entirely omitting IDP and CFP modules.  ``w/o \normalsize{\textcircled{\scriptsize{2}}}\normalsize \normalsize{\textcircled{\scriptsize{3}}}'' indicates training without module IDP and inference with a fixed signal. ``w/o \normalsize{\textcircled{\scriptsize{2}}}'' refers to using a fixed inference signal.  ``w/o \normalsize{\textcircled{\scriptsize{3}}}'' indicates training without the IDP module. ``TC'' denotes temporal consistency.}
      \small
       \setlength{\tabcolsep}{1mm}
        \resizebox{0.46\textwidth}{!}{
    \begin{tabular}{l|cccccc}
    \toprule
    Metrics & \multicolumn{3}{c}{Video Quality} & \multicolumn{1}{c}{Alignment} & \multicolumn{2}{c}{Preservation} \\
 
    Model & {TC}$\uparrow$& {FVD}$\downarrow$ & {PSNR}$\uparrow$ & {CLIP-T}$\uparrow$ & {PSNR$_\text{P}$}$\uparrow$ & {SSIM$_\text{P}$}$\uparrow$ \\
    \midrule
    w/o \normalsize{\textcircled{\scriptsize{1}}}\normalsize \normalsize{\textcircled{\scriptsize{2}}}\normalsize \normalsize{\textcircled{\scriptsize{3}}}\normalsize  & \textbf{0.9983} & 621.6  & 20.11  & \underline{12.50} & 26.66  & 0.8873 \\
   w/o \normalsize{\textcircled{\scriptsize{2}}}\normalsize \normalsize{\textcircled{\scriptsize{3}}}\normalsize  & 0.9976 & \underline{80.89} & 26.03  & \textbf{12.27} & 33.75  & \textbf{0.9469} \\
   w/o \normalsize{\textcircled{\scriptsize{2}}}    & \underline{0.9979} & 222.7  & 25.70  & 12.20  & \underline{33.85} & \underline{0.9466} \\
    w/o \normalsize{\textcircled{\scriptsize{3}}}   & 0.9977 & 90.98  & \underline{26.09} & 12.16  & 33.76  & \textbf{0.9469} \\
     \cellcolor[rgb]{ .91,  .91,  .91}O-DisCo-Edit   & \cellcolor[rgb]{ .91,  .91,  .91}0.9978 & \cellcolor[rgb]{ .91,  .91,  .91}\textbf{77.03} & \cellcolor[rgb]{ .91,  .91,  .91}\textbf{26.43} & \cellcolor[rgb]{ .91,  .91,  .91}12.19  & \cellcolor[rgb]{ .91,  .91,  .91}\textbf{33.87} & \cellcolor[rgb]{ .91,  .91,  .91}\underline{0.9466} \\
    \bottomrule
    \end{tabular}}%
\label{tab:model_ablation}
\vspace{-1em}
\end{table}
\begin{figure}[!t]
\centering
\includegraphics[width=0.48\textwidth]{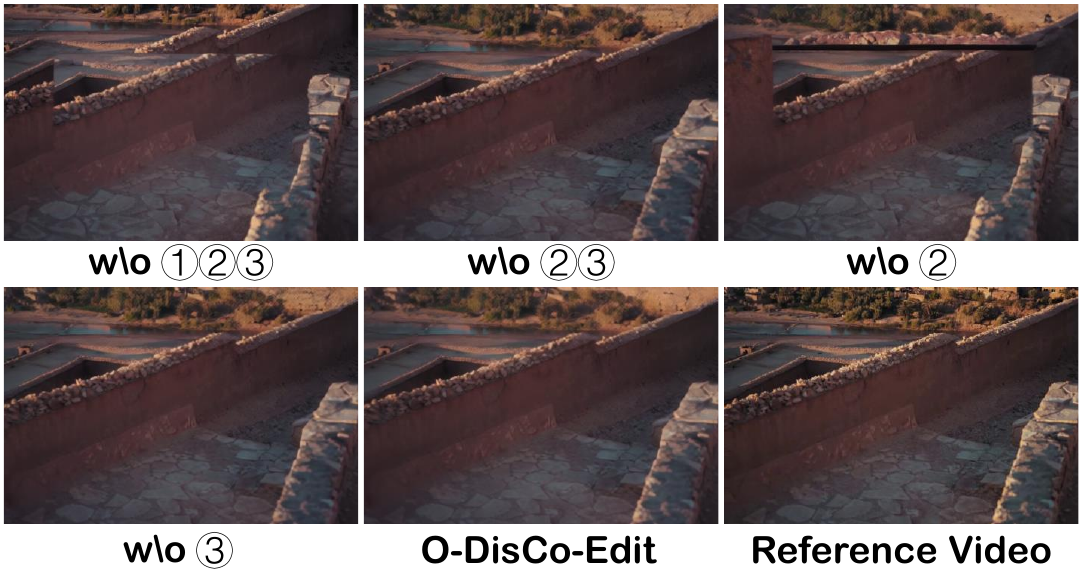}
\caption{Ablation study results for the outpainting task. \normalsize{\textcircled{\scriptsize{1}}}  CFP module,~\normalsize{\textcircled{\scriptsize{2}}}~A-O-DisCo,~\normalsize{\textcircled{\scriptsize{3}}} IDP module. 
    ``w/o \normalsize{\textcircled{\scriptsize{1}}}\normalsize \normalsize{\textcircled{\scriptsize{2}}}\normalsize \normalsize{\textcircled{\scriptsize{3}}}'' denotes training with R-O-DisCo and inference with a fixed signal, entirely omitting IDP and CFP modules.  ``w/o \normalsize{\textcircled{\scriptsize{2}}}\normalsize \normalsize{\textcircled{\scriptsize{3}}}'' indicates training without module IDP and inference with a fixed signal. ``w/o \normalsize{\textcircled{\scriptsize{2}}}'' refers to using a fixed inference signal.  ``w/o \normalsize{\textcircled{\scriptsize{3}}}'' indicates training without the IDP module.}
\vspace{-1em}
\label{fig:outpainting_ablation}
\end{figure}

The ablation study results for the outpainting task are shown in~\cref{tab:model_ablation}. (1) A comparison between row 1 and row 2 reveals that adding the CFP module significantly improves the model's ability to preserve non-edited regions, as evidenced by a substantial increase in both PSNR$_\text{P}$ and SSIM$_\text{P}$. 
(2) A comparison between rows 2 and 3 shows that adding the IDP module without A-O-DisCo results in a significant performance drop. This is because the IDP module emphasizes the position of the edited regions and the corresponding content generation. Therefore, when A-O-DisCo is not introduced, the model is prone to retain the outpainting's boundary information, leading to residual boundary artifacts in the generated video, as illustrated in~\cref{fig:outpainting_ablation}.

\subsection{More Examples}

From~\cref{fig:removal} to~\cref{fig:style}, we present additional qualitative results for tasks including object removal, outpainting, object internal motion transfer, lighting transfer, color change, swap, addition, and style transfer. We extend the object removal task to the DAVIS dataset~\citep{pont20172017}, as shown in~\cref{fig:removal} and~\cref{fig:removal1}. O-DisCo-Edit achieves excellent performance on the DAVIS dataset, demonstrating its effectiveness. Additionally, for the style transfer task, we incorporate VideoComposer~\citep{wang2023videocomposer} as a complementary baseline. The results in~\cref{fig:style} clearly show that O-DisCo-Edit surpasses VideoComposer in terms of style transfer performance.

\section{Limitation}
Our work has several limitations. First, while we compare our model's performance with specialized models in the object removal task, we only compare it with a multi-task model in other tasks. This limited comparison prevents us from presenting a more comprehensive evaluation. Second, we conducted ablation studies for only three tasks, which does not provide a complete understanding of the contributions of different modules to other tasks. Third, the performance of our model heavily depends on the quality of the first frame edit. Poor quality in the first frame edit may result in a significant drop in the model’s performance.

\section{Ethical Consideration}
From an ethical perspective, our model’s powerful video editing capabilities provide creators with innovative tools that can inspire new ideas and enhance the artistic and creative aspects of video content. However, this also raises concerns about the potential for spreading misinformation and false content, which could undermine public trust in information. Additionally, these technologies may unintentionally reinforce existing biases and stereotypes, potentially influencing societal cultural perspectives in a negative way. These issues highlight the importance of ethical reflection and responsibility, urging policymakers, developers, and societal stakeholders to collaboratively establish appropriate regulations to ensure the healthy development of such technologies. In the future, we will make our model publicly available. This release will be accompanied by a licensing agreement that requires users to adhere to our guidelines and outlines acceptable use cases, thereby limiting potential abuse by malicious users.

\begin{figure*}[!t]
\centering
\includegraphics[width=1\textwidth]{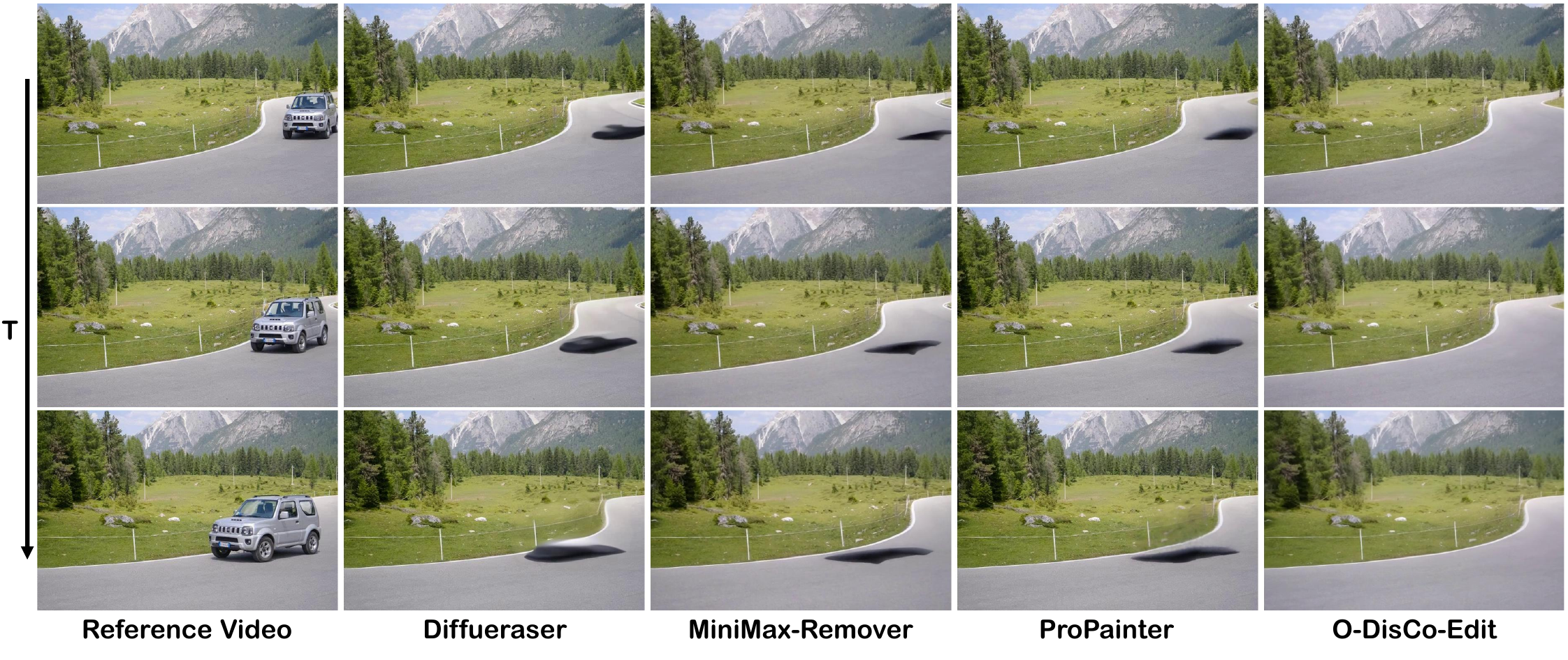}
\caption{Visual comparison of object removal capabilities, benchmarking O-DisCo-Edit against specialized baselines on the DAVIS dataset. ``T'' indicates the temporal axis.}
\vspace{-1em}
\label{fig:removal}
\end{figure*}

\begin{figure*}[!t]
\centering
\includegraphics[width=1\textwidth]{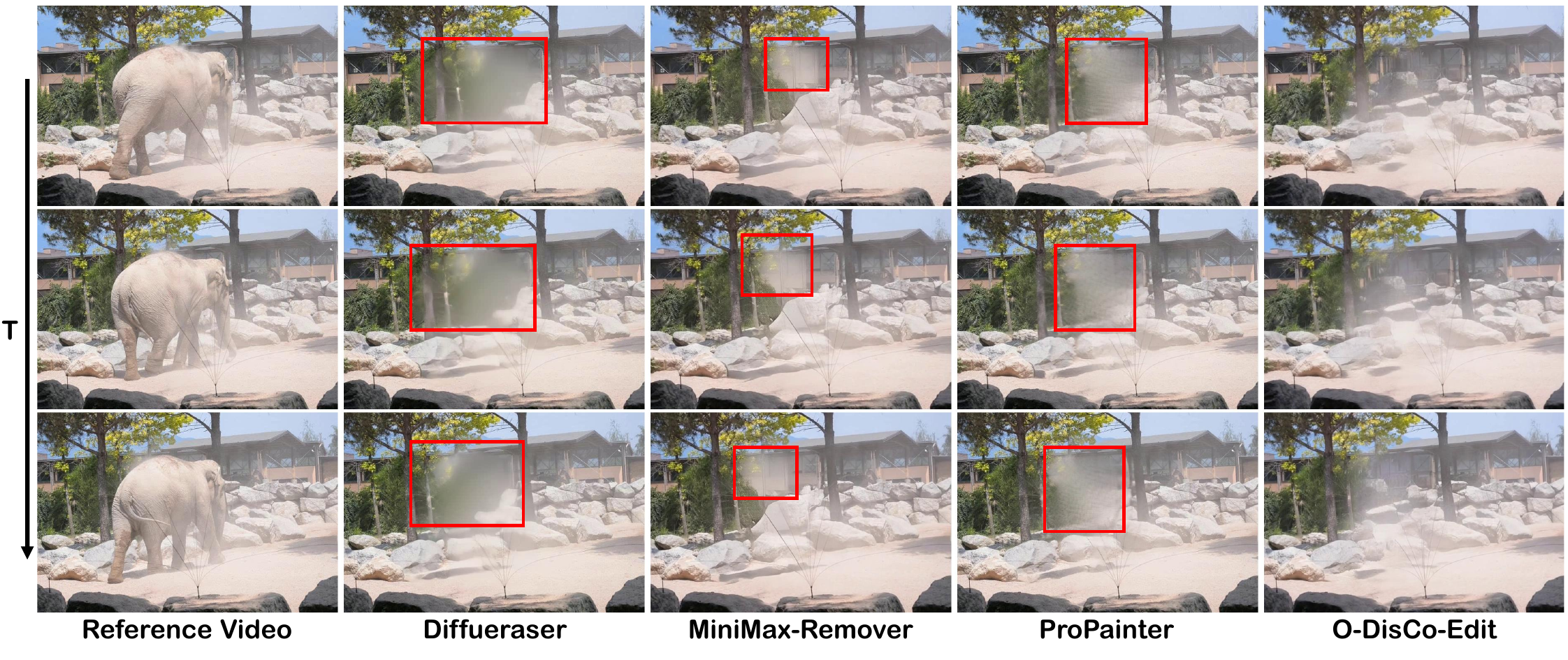}
\caption{Qualitative results for object removal, demonstrating the performance of O-DisCo-Edit versus specialized baselines. The temporal progression is shown along axis ``T''.}
\vspace{-1em}
\label{fig:removal1}
\end{figure*}

\begin{figure*}[!t]
\centering
\includegraphics[width=1\textwidth]{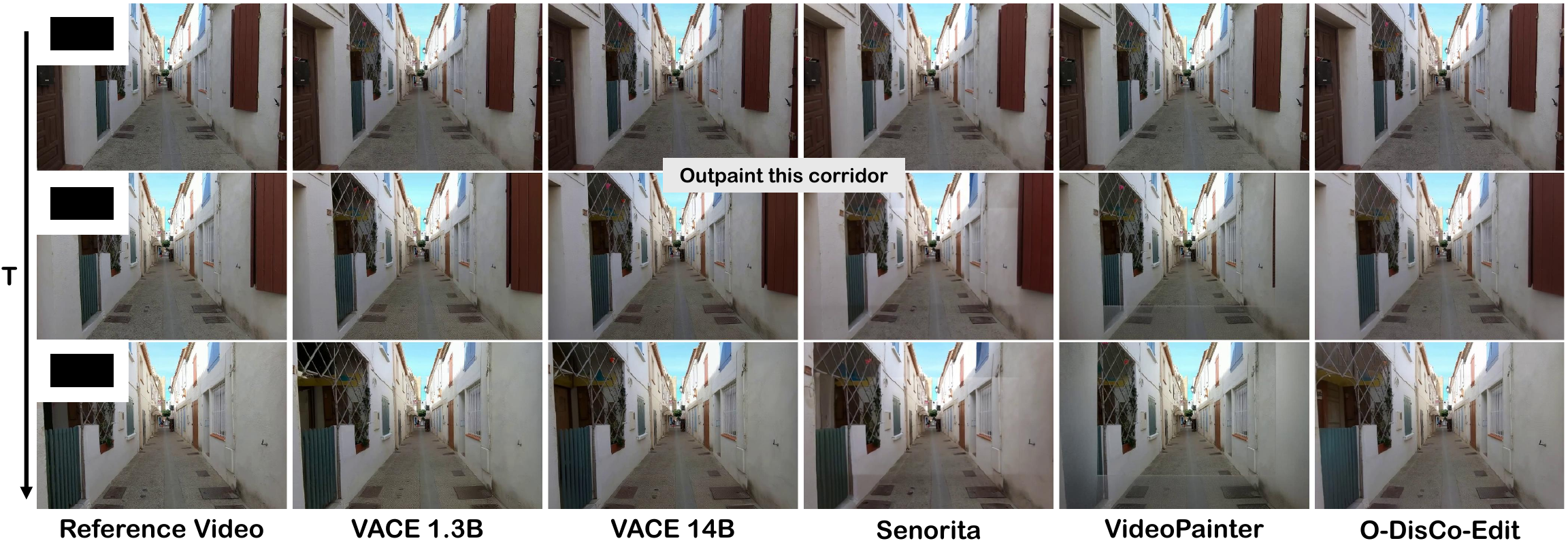}
\caption{Performance visualization for the outpainting task, illustrating how O-DisCo-Edit compares with multi-task baselines on our benchmark. The temporal axis is denoted by ``T''.}
\vspace{-1em}
\label{fig:outpainting}
\end{figure*}

\begin{figure*}[!t]
\centering
\includegraphics[width=1\textwidth]{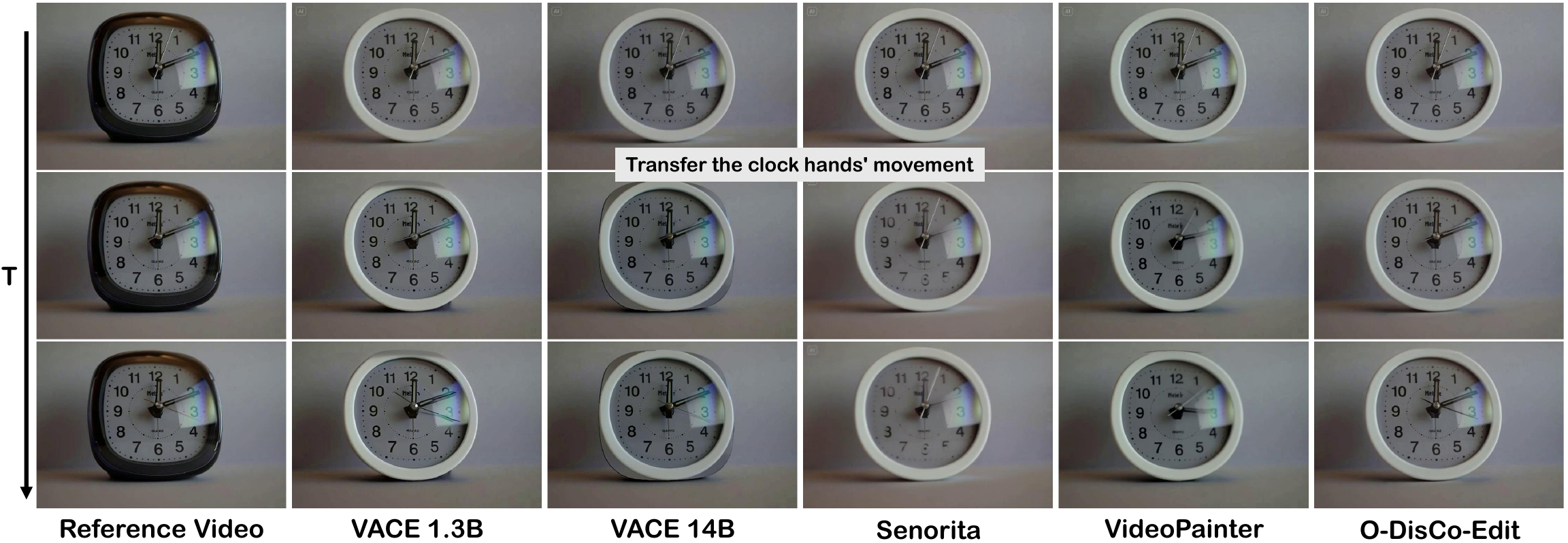}
\caption{A comparative evaluation of object internal transfer performance between O-DisCo-Edit and mutil-task baselines on the our benchmark, where ``T'' denotes the temporal axis.}
\label{fig:Tranfer_move}
\end{figure*}

\begin{figure*}[!t]
\centering
\includegraphics[width=1\textwidth]{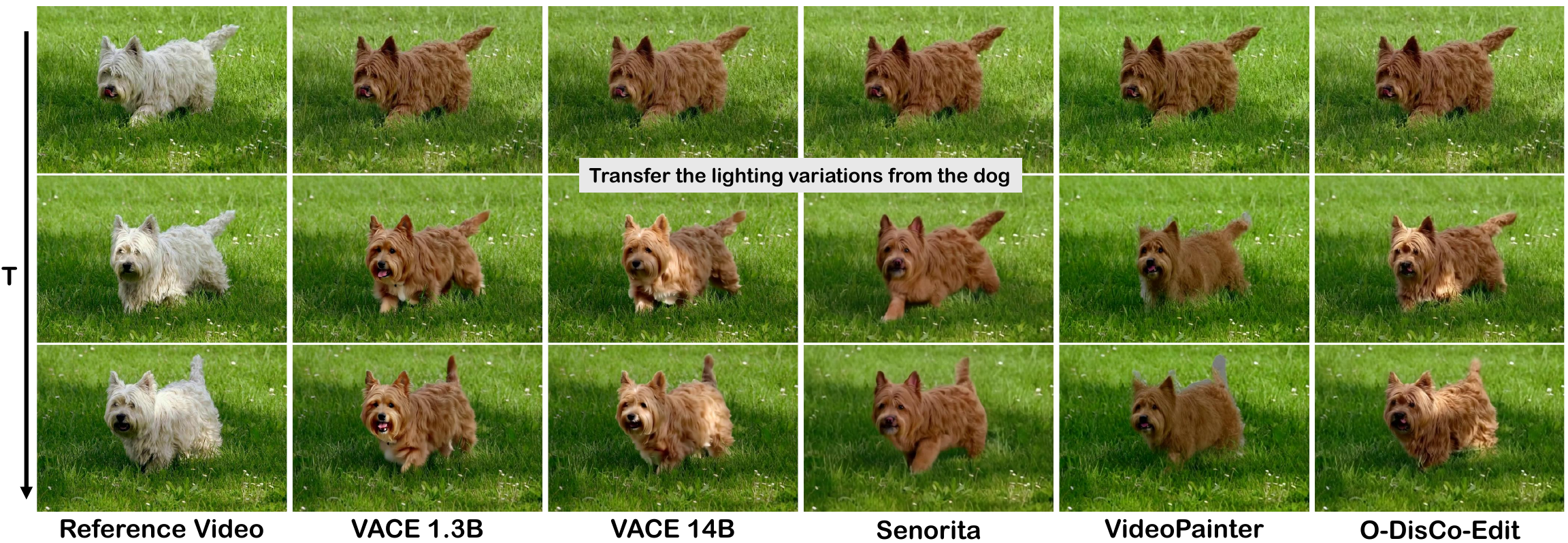}
\caption{A comparison of lighting transfer performance between O-DisCo-Edit and multi-task baselines on our benchmark. The temporal progression is marked by ``T''.}
\label{fig:Tranfer_light}
\end{figure*}

\begin{figure*}[!t]
\centering
\includegraphics[width=1\textwidth]{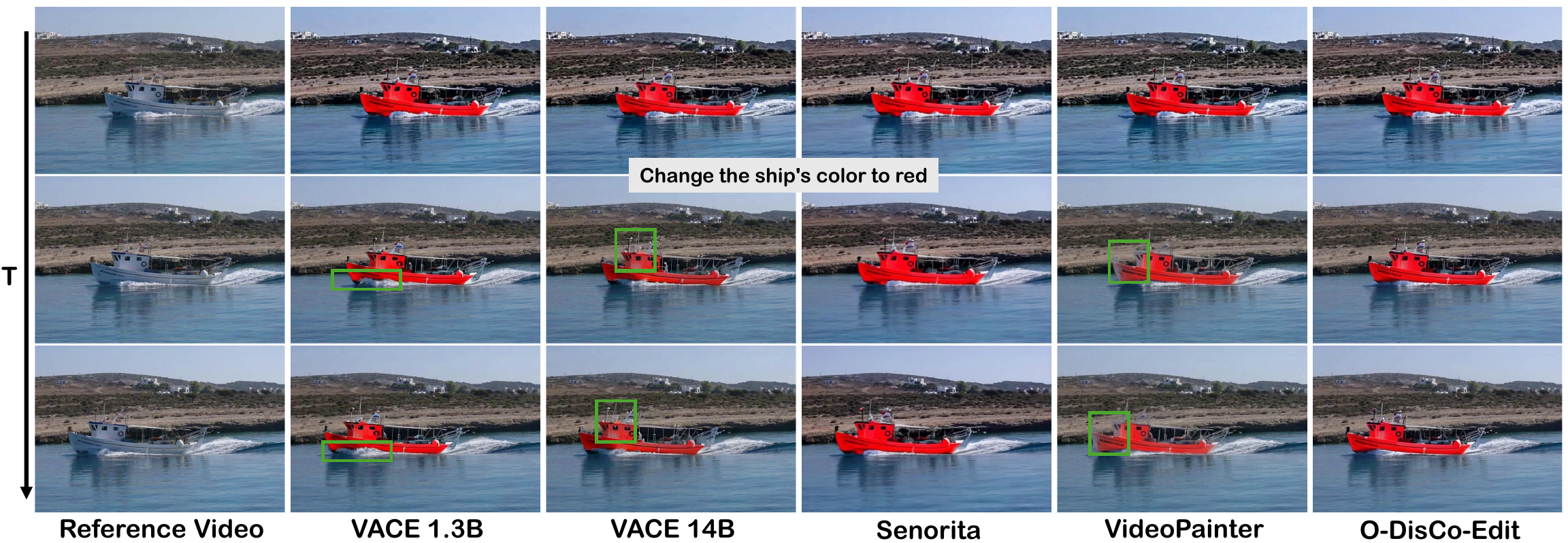}
\caption{A comparative view of color change performance between O-DisCo-Edit and multi-task baselines on our benchmark. The time axis is indicated by 'T'.}
\label{fig:Color_change}
\end{figure*}

\begin{figure*}[!t]
\centering
\includegraphics[width=1\textwidth]{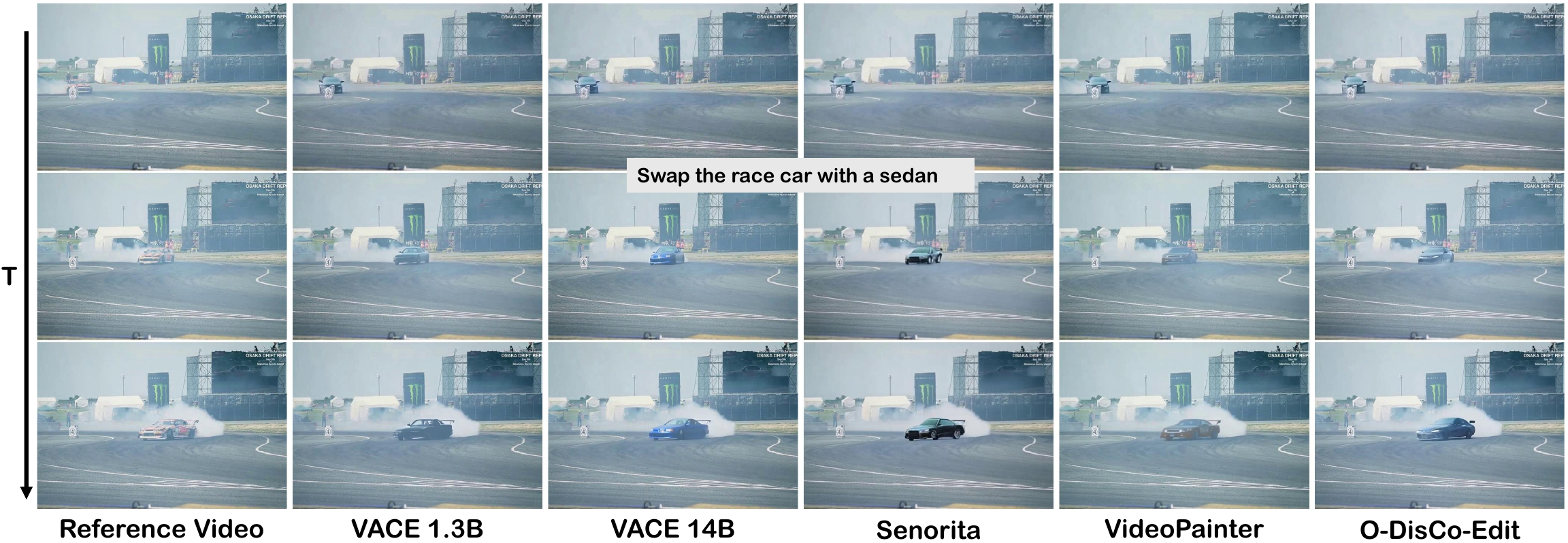}
\caption{Qualitative results for the object swap task, comparing the output of O-DisCo-Edit against multi-task baselines on our benchmark. ``T'' denotes the temporal axis.}
\label{fig:Swap}

\end{figure*}
\begin{figure*}[!t]
\centering
\includegraphics[width=1\textwidth]{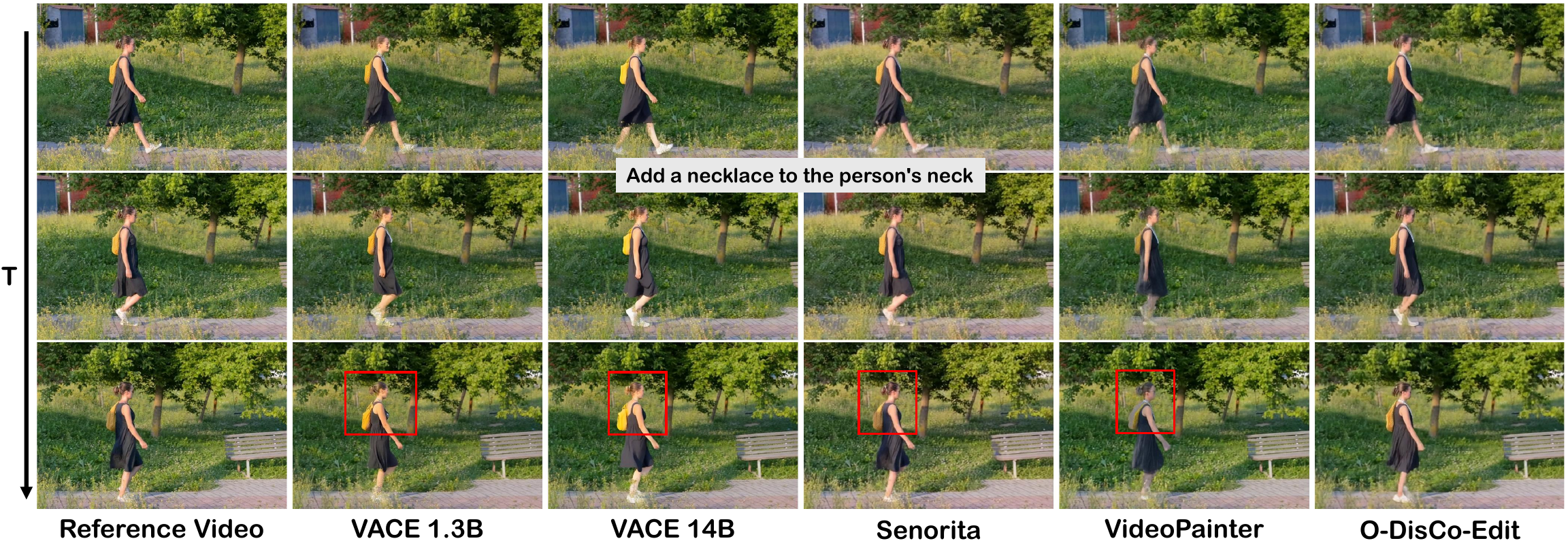}
\caption{An evaluation of addtion performance, highlighting the differences between O-DisCo-Edit and multi-task baselines on our benchmark. The temporal axis is shown as ``T''.}
\label{fig:addtion}
\end{figure*}

\begin{figure*}[!t]
\centering
\includegraphics[width=1\textwidth]{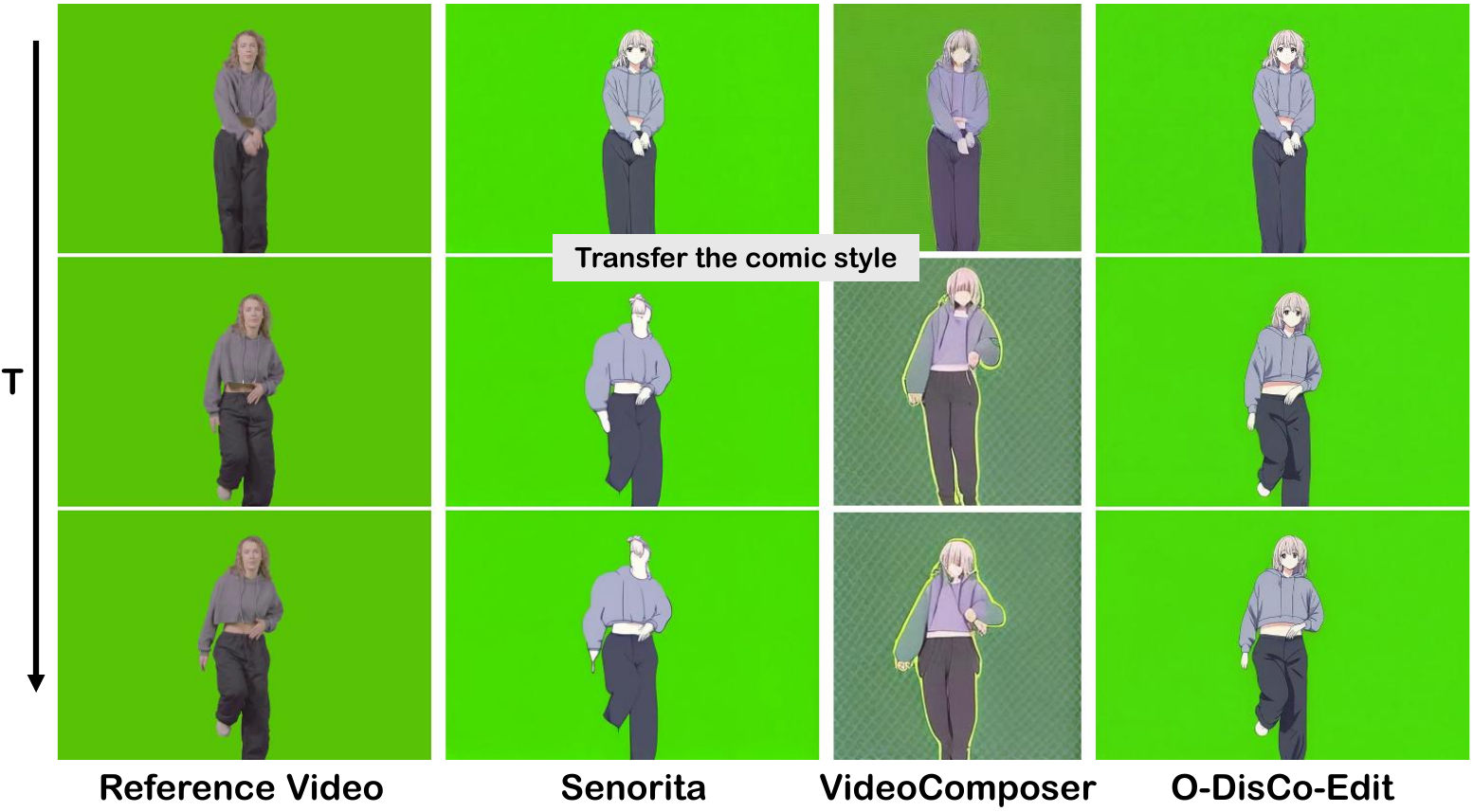}
\caption{A side-by-side comparison of style transfer performance, showcasing O-DisCo-Edit against multi-task baselines on our benchmark. The letter ``T'' denotes the temporal dimension.}
\label{fig:style}
\end{figure*}

\clearpage

\end{document}